%% file: jmlr_draft_arxiv.tex
\documentclass[twoside,11pt]{article}

\usepackage{jmlr2e}
\usepackage{amsmath,amsfonts,ifthen,verbatim, amssymb, color,mathrsfs,eucal}
\usepackage{graphicx,appendix,framed,float}
\usepackage[algoruled,linesnumbered]{algorithm2e}
\usepackage{setspace,multirow,textcomp}

\usepackage[all]{xy}

\jmlrheading{???}{???}{???}{???}{???}{Mohammad Gheshlaghi Azar Vicen\c{c} G\'omez and Hilbert J. Kappen}

\long\def\symbolfootnote[#1]#2{\begingroup
\def\thefootnote{\fnsymbol{footnote}}\footnote[#1]{#2}\endgroup}

\ShortHeadings{Dynamic Policy Programming}{Gheshlaghi Azar, G\'omez and Kappen}

\usepackage{color,soul}

\begin{document}



\title{Dynamic Policy Programming}
\author{\name Mohammad Gheshlaghi Azar \email m.azar@science.ru.nl \\
\name Vicen\c{c} G\'omez
\email v.gomez@science.ru.nl
\\
\name Hilbert J. Kappen
\email b.kappen@science.ru.nl \\
\addr Department of Biophysics\\ Radboud University Nijmegen\\ 6525 EZ Nijmegen, The Netherlands\\
}
\editor{TBA}
\maketitle 

\newtheorem{prop}{Proposition}
\newtheorem{defn}{Definition}
\newtheorem{asm}{Assumption}


\newcommand{\cl}[2][ (]{
\ifthenelse{\equal{#1}{ (}}{\left (#2\right)}{}
\ifthenelse{\equal{#1}{[}}{\left[#2\right]}{}
\ifthenelse{\equal{#1}{\{}}{\left\{#2\right\}}{}
}

\newcommand{\inset}[3][C]{
\ifthenelse{\equal{#1}{C}}{{#2}\in\mathcal{#3}}{}
\ifthenelse{\equal{#1}{N}}{{#2}\in{#3}}{}
}

\newcommand{\pol}[1][N]{
\ifthenelse{\equal{#1}{N}}{\pi(a|x)}{}
\ifthenelse{\equal{#1}{T}}{\pi{(\cdot|x)}}{}
\ifthenelse{\equal{#1}{X}}{\Pi_{x}}{}
\ifthenelse{\equal{#1}{-}}{\bar{\pi}(a|x)}{}
\ifthenelse{\equal{#1}{T-}}{\bar{\pi}(\cdot|x)}{}
\ifthenelse{\equal{#1}{S-}}{\bar{\Pi}_{x}}{}
}

\newcommand{\Ts}{T_{xx'}^{a}}

\newcommand{\subExt}[3][max]{
\ifthenelse{\equal{#1}{max}}{\underset{\inset[N]{#2}{#3}}{\max}}{}
\ifthenelse{\equal{#1}{sup}}{\underset{\inset[N]{#2}{#3}}{\sup}}{}
\ifthenelse{\equal{#1}{inf}}{\underset{\inset[N]{#2}{#3}}{\sup}}{}
\ifthenelse{\equal{#1}{min}}{\underset{\inset[N]{#2}{#3}}{\min}}{}
\ifthenelse{\equal{#1}{maxC}}{\underset{\inset{#2}{#3}}{\max}}{}
\ifthenelse{\equal{#1}{supC}}{\underset{\inset{#2}{#3}}{\sup}}{}
\ifthenelse{\equal{#1}{infC}}{\underset{\inset{#2}{#3}}{\sup}}{}
\ifthenelse{\equal{#1}{minC}}{\underset{\inset{#2}{#3}}{\min}}{}
}

\newcommand{\ESum}[4][C]
{\ifthenelse{\equal{#1}{C}}{\underset{\inset{#3}{#4}}{\sum}#2}{}
 \ifthenelse{\equal{#1}{N}}{\underset{\inset[N]{#3}{#4}}{\sum}#2}{}
\ifthenelse{\equal{#1}{U}}{\overset{#4}{\underset{#3}{\sum}}#2}{}
\ifthenelse{\equal{#1}{X}}{\sideset{}{_{#3}^{#4}}{\sum}#2}{}
\ifthenelse{\equal{#1}{S}}{\sideset{}{_{#3}^{#4}}{\sum}#2}{}
\ifthenelse{\equal{#1}{O}}{{\underset{ (#3,#4)}{\sum}}#2}{}
\ifthenelse{\equal{#1}{I}}{{\underset{#3}{\sum}}#2}{}}

\newcommand{\VF}[2][N]{
\ifthenelse{\equal{#1}{L}}{V^{\pi}_{\eta} (#2)}{}
\ifthenelse{\equal{#1}{C}}{V{^{\pi} (#2)}}{}
\ifthenelse{\equal{#1}{N}}{V^*_{\bar{\pi}} (#2)}{}
\ifthenelse{\equal{#1}{CO}}{V{^{*} (#2)}}{}
\ifthenelse{\equal{#1}{Mx}}{V^{\pi}_{\infty} (#2)}{}
\ifthenelse{\equal{#1}{Mxo}}{V^{\pi^*}_{\infty} (#2)}{}
\ifthenelse{\equal{#1}{Mn}}{V^{\pi}_{-\infty} (#2)}{}
\ifthenelse{\equal{#1}{Mno}}{V^{\pi^*}_{-\infty} (#2)}{}
}
\newcommand{\CE}[4]{\ESum{#1\log\cl{\frac{#1}{#2}}}{#3}{#4}}
\newcommand{\KL}[2]{\mathrm{KL}\cl{#1\|#2}}
\newcommand{\Eval}[1][null]{
\ifthenelse{\equal{#1}{null}}{\mathbb{E}}{\mathbb{E}_{#1}}
}
\newcommand{\ELog}[1][null]{
\ifthenelse{\equal{#1}{null}}{\mathbb{E}^{\eta}}{\mathbb{E}^{\eta}_{#1}}
}

\newcommand{\adit}[1]{\textit{\textbf{#1}}}
\newcommand{\inv}[2][frac]{
\ifthenelse{\equal{#1}{frac}}{\frac{1}{#2}}{}
\ifthenelse{\equal{#1}{inv}}{{#2}^{-1}}{}
}

\newcommand{\pdiff}[2]{
\frac{\partial{#1}}{\partial{#2}}
}

\newcommand{\M}[1][\eta]{
\mathcal{M}_{#1}
}

\newcommand{\Z}[1][\eta]{
\mathcal{Z}_{#1}
}

\newcommand{\OP}[1][null]{
\ifthenelse{\equal{#1}{null}}{\mathcal{O}}{\mathcal{O}^{#1}}
}

\newcommand{\pv}{{\pi}}

\newcommand{\qv}[1][null]{
\ifthenelse{\equal{#1}{null}}{Q^*}{Q^{#1}}
}
\newcommand{\OPM}[2][null]{
\ifthenelse{\equal{#1}{null}}{{\mathcal{O}}{#2}}{\ifthenelse{\equal{#1}{N}}{{\mathcal{O}}{#2}}{{\mathcal{O}}^{#1}{#2}}}
}

\newcommand{\OPS}[2][null]{
\ifthenelse{\equal{#1}{null}}{{#2}_{#1}}{{#2}_{#1}}
}

\newcommand{\OPMS}[2][null]{
\ifthenelse{\equal{#1}{null}}{{#2}_{#1}}{{#2}_{#1}}
}

\newcommand{\U}[1][null]{
\ifthenelse{\equal{#1}{null}}{\mathcal{U}_{\eta}}{\mathcal{U}_{\eta}^{#1}}
}

\newcommand{\T}[1][null]{
\ifthenelse{\equal{#1}{null}}{\mathcal{T}}{\mathcal{T}^{#1}}
}

\newcommand{\F}[1][null]{
\ifthenelse{\equal{#1}{null}}{\mathcal{F}}{\mathcal{F}^{#1}}
}

\newcommand{\G}[1][null]{
\ifthenelse{\equal{#1}{null}}{\mathcal{G}}{\mathcal{G}^{#1}}
}

\newcommand{\subLim}[2]{
\underset{#1\rightarrow#2}{\lim}
}

\newcommand{\Norm}[2][]
{
\left\|#2\right\|_{#1}
}

\newcommand{\bldsym}{\boldsymbol}

\newcommand{\TM}[1][null]{
\ifthenelse{\equal{#1}{null}}{{\mathcal{T}}}{{\mathcal{T}}^{#1}}
}

\newcommand{\rv}{r}

\newcommand{\MM}{
\mathcal{M}
}

\definecolor{darkblue}{rgb}{0,0.9,0} 
\sethlcolor{darkblue}


\begin{abstract}
In this paper, we propose a novel  policy iteration method, called dynamic policy programming (DPP), to estimate the optimal policy in the infinite-horizon Markov decision processes.  We  prove the finite-iteration and asymptotic  $\ell_{\infty}$-norm performance-loss bounds  for DPP in the presence of approximation/estimation error. 
The bounds are expressed in terms of  the $\ell_{\infty}$-norm of the average accumulated error as opposed to the $\ell_{\infty}$-norm of the error in the case of the standard approximate value iteration (AVI) and the approximate policy iteration (API). This suggests that  DPP can achieve a better performance than AVI and API since it ~\emph{averages out} the simulation noise caused by Monte-Carlo sampling throughout the learning process.  We examine this theoretical results numerically by comparing the  performance of the approximate variants of DPP with  existing reinforcement learning (RL)  methods on different problem domains. Our results  show that, in all cases, DPP-based algorithms outperform other RL methods by a wide margin. 
\end{abstract}

\begin{keywords}
Approximate dynamic programming, reinforcement learning, Markov decision processes, Monte-Carlo methods, function approximation.
\end{keywords}




\section{Introduction}
\label{intro}
Many problems in robotics, operations research and process control can be
represented as a control problem that can be solved by 
finding the optimal policy using \emph{dynamic programming} (DP).
DP is based on the estimating some measures of the value of  state-action $Q^*(x,a)$ through the Bellman equation.
  For high-dimensional discrete systems or for continuous systems, computing the value function by DP is
intractable. The common approach to make the computation tractable is to
approximate the value function using function-approximation and Monte-Carlo
sampling~\citep{Szepesvari2010,Tsitsiklis96}.   Examples of such approximate
dynamic programming (ADP) methods are approximate policy iteration (API) and
approximate value iteration (AVI)~\citep{Berst07b,Lago03,PerkinsP02,Farias00}. 

ADP methods have been successfully applied to many real world problems, and
theoretical results have been derived in the form of finite iteration and asymptotic
performance guarantee of the induced policy~\citep{AmirRemi10,Thiery10,Remi05,Tsitsiklis96}.
The asymptotic $\ell_{\infty}$-norm performance-loss bounds of API and AVI are expressed in terms of the supremum, with respect to (w.r.t.) the number of iterations, of the approximation errors:
\begin{equation*}
\underset{k\rightarrow\infty}{{\lim\sup}}\|Q^*-Q^{\pi_k}\|\leq\frac{2\gamma}{(1-\gamma)^2}\underset{k\rightarrow\infty}{{\lim\sup}}\Norm{\epsilon_k},
\end{equation*}
where $\gamma$ denotes the discount factor, $\|\cdot\|$ is the $\ell_{\infty}$-norm  w.r.t. the state-action pair $(x,a)$. Also,  $\pi_k$ and $\epsilon_k$ are the control policy and the approximation error at  round $k$ of the ADP algorithms, respectively.  In many problems of interest, however, the supremum over the normed-error $\Norm{\epsilon_k}$  can be  large and hard to control due to the large variance of  estimation caused by Monte-Carlo sampling. In those cases,  a bound which instead depends  on  the average accumulated  error $\bar{\epsilon}_k=1/(k+1)\sum{_{j=0}^{k}}{\epsilon_j}$ is preferable. This is due to the fact that the  errors associated with the variance of estimation can be considered as the instances of some zero-mean random variables. Therefore, one can show, by making use of  a law of large numbers argument,  that those errors are asymptotically \emph{averaged out} by accumulating the approximation errors of all iterations.\footnote{The law of large numbers requires the errors to satisfy some  stochastic assumptions, e.g., they need to be identically and independently distributed (i.i.d.) samples or  martingale differences.}

In this paper, we propose a new mathematically-justified approach to estimate the optimal policy, called
dynamic policy programming (DPP).  We prove finite-iteration and asymptotic performance loss bounds for the policy induced by DPP in the presence of approximation. The  asymptotic bound of approximate DPP is   expressed in terms of the average accumulated error $\|\bar{\epsilon}_k\|$ as opposed to $\|\epsilon_k\|$ in the case of AVI and API.  This result suggests that DPP may perform better than AVI and API   in the presence of large variance of estimation since it can average out the estimation errors throughout the learning process. The dependency on the average error $\|\bar{\epsilon}_k\|$ follows naturally from the incremental policy update of DPP which at each round of policy update, unlike AVI and API, accumulates the approximation errors of the previous iterations, rather than just minimizing the approximation error of the current iteration.

This article is organized as follows.  In Section~\ref{Prelim}, we present the
notations which are used in this paper.   We introduce DPP and we investigate
its  convergence properties in Section~\ref{Alg}.   In Section~\ref{Approx}, we
demonstrate the compatibility of our method with the approximation techniques.  We generalize DPP bounds to the case
of function approximation and Monte-Carlo simulation. We also introduce a new convergent  RL algorithm, called DPP-RL, which 
relies on an approximate sample-based  variant of DPP to estimate the optimal policy.   
Section~\ref{NumRes}, presents numerical experiments on several problem domains including the optimal
replacement problem~\citep{Munos08} and a stochastic grid world.
In Section~\ref{rel} we briefly review some related work.
 Finally, we discuss some of the implications of our work in Section~\ref{conclus}. 

\section{Preliminaries}
\label{Prelim}

In this section, we introduce some concepts and definitions from the theory of Markov decision processes (MDPs) and reinforcement learning (RL) as well as some standard notations.\footnote{For further reading see~\citet{Szepesvari2010}.} We begin by  the definition of  the $\ell_{2}$-norm (Euclidean norm) and the $\ell_{\infty}$-norm (supremum norm). Assume that $\mathcal{Y}$ is a finite set. Given the probability measure $\mu$ over $\mathcal{Y}$, for a real-valued function $g:\mathcal{Y}\to{\mathbb{R}}$, we shall denote the $\ell_{2}$-norm and the weighted  $\ell_{2,\mu}$-norm of $g$ by $\Norm[2]{g}^2\triangleq\sum_{\inset{y}{Y}}{g(y)^2}$  and  $\Norm[2,\mu]{g}^2\triangleq\sum_{\inset{y}{Y}}{\mu(y)g(y)^2}$, respectively. Also, the $\ell_{\infty}$-norm of  $g$ is  defined by $\Norm{g}\triangleq\max_{\inset{y}{Y}}|g(y)|$.
  
\subsection{Markov Decision Processes}
A discounted MDP is a quintuple $\cl{\mathcal{X,A},P,\mathcal{R},\gamma}$, where $\mathcal{X}$ and $\mathcal{A}$ are, respectively, the state space and the action space. $P$ shall denote the state transition distribution  and $\mathcal{R}$ denotes the reward kernel. $\gamma\in[0,1)$ denotes the discount factor. The transition  $P$ is a probability kernel over the next state upon taking action $a$  from state $x$, which we shall denote by $P(\cdot|x,a)$. $\mathcal{R}$ is a set of real-valued numbers. A reward $r(x,a)\in\mathcal{R}$ is associated with each state $x$ and action $a$.  To keep the representation succinct, we shall denote the joint state-action space $\mathcal{X}\times\mathcal{A}$ by  $\mathcal{Z}$.

\begin{asm}[MDP Regularity]
\label{Regular}
We assume $\mathcal{X}$ and $\mathcal{A}=\{a_1,a_2,\dots
,a_L\}$ are finite sets. Also,  the absolute value of the immediate reward $r(x,a)$ is bounded from above by $R_{\max}>0$ for all $\inset{(x,a)}{Z}$. We also define $V_{\max}=R_{\max}\big/(1-\gamma)$.
 \end{asm}           
A policy kernel $\pi(\cdot|\cdot)$ determines  the distribution of the control action given the past observations. The policy is called stationary and Markovian if the distribution of the control action is independent of time and only depends on the last state $x$.  Given the last state $x$, we shall denote the stationary policy by $\pi(\cdot|x)$.   A stationary policy is called deterministic if for any state $x$ there exists some  action $a$ such that $\pi(\cdot|x)$ concentrates on this action.    Given the policy $\pi$ its corresponding value function $V^{\pi}:\mathcal{X}\to{\mathbb{R}}$ denotes the expected value of the long-term discounted sum of rewards in each state $x$, when the action is chosen by policy $\pi$ which we denote by $V^{\pi}(x)$.  Often it is convenient to associate value functions not with states but with state-action pairs. Therefore, we introduce $Q^{\pi}:\mathcal{Z}\to\mathbb{R}$ as the expected total discounted reward upon choosing action $a$ from state $x$ and then following policy $\pi$, which we shall denote by $Q^{\pi}(x,a)$. We define the \emph{Bellman operator} $\T[\pi]$ on the action-value functions by:
\begin{equation*}
\begin{aligned}
\T[\pi]Q(x,a)\triangleq{r}(x,a)+\gamma{\sum_{\inset{(x',a')}{Z}}P(x'|x,a)\pi(a'|x')Q(x',a')},&&&&\forall\inset{(x,a)}{Z}.
\end{aligned}
\end{equation*}

We also notice that $Q^{\pi}$ is the fixed point of $\T[\pi]$.

The goal is to find a policy $\pi^*$ that attains the \emph{optimal value function}, $V^*(x)\triangleq\sup_{\pi}V^{\pi}(x)$, at all states $\inset{x}{X}$. The optimal value function satisfies the Bellman equation:
\begin{align}
\label{ValBellmanStoch}
\begin{aligned}
V^*(x)&=\sup_{\pi}\sum_{\substack{\inset{x'}{X}\\\inset{a}{A}}}\pi(a|x)\left[r(x,a)+P(x'|x,a)V^*(x')\right]
\\&=\max_{\inset{a}{A}}\left[r(x,a)+\sum_{\inset{x'}{X}}{P(x'|x,a)V^*(x')}\right]
\end{aligned},&&&&\forall\inset{x}{X}.
\end{align}

Likewise, the \emph{optimal action-value function} $Q^*$ is defined by  $Q^*(x,a)=\sup_{\pi}Q^{\pi}(x,a)$ for all $\inset{(x,a)}{Z}$. We shall define the \emph{Bellman optimality operator}  $\mathcal{T}$ on the action-value functions as:
\begin{equation*}
\begin{aligned}
\mathcal{T}Q(x,a)\triangleq{r}(x,a)+\gamma\sum_{\inset{x'}{X}}{P(x'|x,a)\max_{\inset{a'}{A}}Q(x',a')},&&&&\forall\inset{(x,a)}{Z}.
\end{aligned}
\end{equation*}

Likewise, $Q^*$ is the fixed point of $\T$.
 
 Both $\T$ and $\T[\pi]$ are contraction mappings, w.r.t. the supremum norm, with the factor $\gamma$~\citep[chap.~1]{Berst07b}. In other words, for any  two action-value functions  $Q$ and $Q'$,  we have:
\begin{align}
\label{TCont}
\Norm{\T{Q}-\T{Q}'}\leq\gamma\Norm{Q-Q'},&&\Norm{\T[\pi]Q-\T[\pi]{Q'}}\leq\gamma\Norm{Q-Q'}.
\end{align}
    
The policy distribution $\pi$ defines the state-action transition kernel $P^{\pi}: M(\mathcal{Z})\to{M}(\mathcal{Z})$, where $M$ is the space of all probability measures defined on $\mathcal{Z}$, as:
\begin{equation*}
P^{\pi}(x',a'|x,a)\triangleq\pi(a'|x')P(x'|x,a).
\end{equation*}

From this kernel a right-linear operator $P^{\pi}\cdot$ is defined by:
 \begin{equation*}
 \begin{aligned}
 (P^{\pi}Q)(x,a)\triangleq\sum_{\inset{(x',a')}{Z}}P^{\pi}(x',a'|x,a){Q}(x',a'),&&&&\forall\inset{(x,a)}{Z}.
 \end{aligned}
 \end{equation*}

Further, we  define two other right-linear operators $\pi\cdot$ and $P\cdot$ by:
\begin{equation*}
\begin{aligned}
(\pi{Q})(x)&\triangleq\sum_{\inset{a}{A}}\pi(a|x){Q}(x,a),&&\forall\inset{x}{X},\\
(P{V})(x,a)&\triangleq\ESum{P(x'|x,a){V}(x')}{x'}{X},&&\forall\inset{(x,a)}{Z}.
\end{aligned}
\end{equation*}

We define the max operator $\mathcal{M}$ on the action value functions by $(\mathcal{M}Q)(x)\triangleq\max_{\inset{a}{A}}Q(x,a)$, for all $\inset{x}{X}$.  Based on the new definitions  one can rephrase the Bellman operator and the Bellman optimality operator as:
\begin{equation}
\label{BellmanOP}
\begin{aligned}
\T[\pi]{Q}(x,a)=r(x,a)+\gamma(P^{\pi}Q)(x,a),&&&&\mathcal{T}{Q}(x,a)=r(x,a)+\gamma(P\mathcal{M}Q)(x,a).
\end{aligned}
\end{equation}

        
\section{Dynamic Policy Programming}
\label{Alg}
In this section, we derive the DPP algorithm starting from the Bellman equation.  We first show that by adding a  relative entropy term to the reward we can control the deviations of the induced policy from a baseline policy. 
We then derive an iterative double-loop approach which combines value and policy updates.  We  reduce this double-loop iteration to just a single iteration by introducing DPP algorithm.   We emphasize that the purpose of the following~\textit{derivations} is to motivate
DPP, rather than to provide a formal characterization.   Subsequently, in Subsection~\ref{DPPProof} and Section~\ref{Approx} , we theoretically investigate the finite-iteration and the asymptotic behavior of DPP and prove its convergence. 
\subsection{From Bellman Equation to DPP Recursion} 
\label{sec:fromBellmantoheaven}
Consider the relative entropy between the policy $\pi$ and 
some baseline policy $\bar{\pi}$:
\begin{align*}
\begin{aligned}
g^{\pi}_{\bar{\pi}} (x)&\triangleq\KL{\pi(\cdot|x)}{\bar{\pi}(\cdot|x)}=\ESum{\pi(a|x)\log\cl[[]{\dfrac{\pi(a|x)}{\bar{\pi}(a|x)}}}{a}{A},
\end{aligned}&&\forall\inset{x}{X}. 
\end{align*}
We define a new value function $V^\pi_{\bar{\pi}}$,  for all $\inset{x}{X}$,
which incorporates $g$ as a penalty term for deviating from the base policy $\bar{\pi}$ and the reward under the policy $\pi$:
\begin{equation*}
 V^{\pi}_{\bar{\pi}}(x)
 \triangleq\quad\subLim{n}{\infty}\Eval{\cl[[]{\left. \ESum[U]{\gamma^{k} \left(r_{t+k}-\frac{1}{\eta}g^{\pi}_{\bar{\pi}} (x_{t+k})\right)\right|x_t=x}{k=0}{n}}},
 \end{equation*}
where $\eta$ is  a  positive constant and $r_{t+k}$ is the reward at time $t+k$. Also, the expected value is taken 
w.r.t.  the state transition probability distribution $P$ and the policy $\pi$. 
The optimal value function $\VF{x}\triangleq\sup_{\pi} V^{\pi}_{\bar{\pi}} (x)$ then
satisfies the following Bellman equation for all $\inset{x}{X}$:
\begin{equation} 
\label{ExpValKLMax1}
\VF{x}=\quad\underset{\pi}{\sup}\sum_{\substack{\inset{a}{A}}}\pi(a|x)
\left[r(x,a)-\frac{1}{\eta}\log\frac{\pi(a|x)}{\bar{\pi}(a|x)}+\gamma(PV^*_{\bar{\pi}})(x,a)\right].
\end{equation}
Equation \eqref{ExpValKLMax1} is a modified version of \eqref{ValBellmanStoch}
where, in addition to maximizing the expected reward, the optimal policy
$\bar{\pi}^*$ also minimizes the distance with the baseline policy $\bar{\pi}$.
The maximization in \eqref{ExpValKLMax1} can be performed in closed form.  Following \citet{Todorov06}, we  state  Proposition~\ref{ConstLem}:
\begin{prop}
\label{ConstLem}
Let $\eta$ be a positive constant, then  for all $\inset{x}{X}$ the  optimal value function $\VF{x}$  and   for all $\inset{(x,a)}{Z}$ the optimal policy $\bar{\pi}^*(a|x)$, respectively, satisfy: 
\begin{align}
\label{ExpValBellmanFinal}
\VF{x}&=\frac{1}{\eta}\log\ESum{\bar{\pi}(a|x)\exp\bigl[\eta({r}(x,a)+\gamma{(PV^{*}_{\bar{\pi}})}(x,a))\bigr]}{a}{A},
\\
\label{ExpPolFinal}
\bar{\pi}^*(a|x)&=\dfrac{\bar{\pi}(a|x)\exp\bigl[\eta({r}(x,a)+\gamma{(PV^{*}_{\bar{\pi}})}(x,a))\bigr]}{\exp\cl{\eta\VF{x}}}.
\end{align}
\end{prop}

\begin{proof}
See Appendix~\ref{AppLem1}. 
\end{proof}

The optimal policy $\bar{\pi}^*$ is a function of the base policy, the optimal
value function $V^*_{\bar{\pi}}$ and the state transition probability $P$.   One can first obtain the
optimal value function $V_{\bar{\pi}}^*$ through the following fixed-point
iteration:
\begin{equation}
\label{ValIter}
V^{k+1}_{\bar{\pi}}(x)=\frac{1}{\eta}\log\ESum{\bar{\pi}(a|x)\exp\bigl[\eta({r}(x,a)+\gamma{(PV^{k}_{\bar{\pi}})}(x,a))\bigr]}{a}{A},
\end{equation}
and then compute  $\bar{\pi}^*$ using \eqref{ExpPolFinal}. 
 $\bar{\pi}^*$ maximizes the value function $V^{\pi}_{\bar{\pi}}$.  However, we are not, in principle, interested in quantifying $\bar{\pi}^*$, but in solving the original MDP problem and computing $\pi^*$. The idea to further improve the policy towards $\pi^*$  is to replace the base-line policy with the just newly computed policy of~\eqref{ExpPolFinal}.  The new policy can be regarded as a \emph{new base-line policy}, and the process
can be repeated again.   This leads to a double-loop algorithm to find the
optimal policy $\pi^*$,
where the outer-loop and the inner-loop would consist
of a policy update, Equation \eqref{ExpPolFinal}, 
and a value function update, Equation \eqref{ValIter}, respectively. 

We then follow the following steps to derive the final DPP algorithm:~\textbf{(i)} We introduce some extra smoothness to the policy update rule by replacing the double-loop  algorithm by direct optimization of both value function and
policy simultaneously using the following fixed point iterations:
\begin{align}
\label{ExpValIter}
V^{k+1}_{\bar{\pi}}(x)&=\frac{1}{\eta}\log\ESum{\bar{\pi}(a|x)\exp\bigl[\eta({r}(x,a)+\gamma{(PV^{k}_{\bar{\pi}})}(x,a))\bigr]}{a}{A},
\\
\label{ExpPolIter}
\bar{\pi}_{k+1}(a|x)
&=\dfrac{\bar{\pi}_{k}(a|x)\exp\bigl[\eta({r}(x,a)+\gamma{(PV^{k}_{\bar{\pi}})}(x,a))\bigr]}{\exp\cl{\eta{V}^{k+1}_{\bar{\pi}}(x)}}.
\end{align} 

Further, \textbf{(ii)} we define the action preference function~ $\Psi_k$ \citep{Sutton98}, for all
$\inset{(x,a)}{Z}$ and $k\geq0$, as follows:
\begin{equation}
\label{PrefDef}
\Psi_{k+1}(x,a)\triangleq\frac{1}{\eta}\log{\bar{\pi}_{k}(a|x})+{r(x,a)+\gamma(P{V}^{k}_{\bar{\pi}})(x,a)}. 
\end{equation}

By comparing \eqref{PrefDef} with \eqref{ExpPolIter} and \eqref{ExpValIter}, we deduce:
\begin{align}
\label{PiPref}
\bar{\pi}_k(a|x)&=\frac{\exp(\eta{\Psi}_{k}(x,a))}{\ESum{\exp(\eta{\Psi}_{k}(x,a'))}{a'}{A}},
\\
\label{VPref}
V^{k}_{\bar{\pi}}(x)&=\frac{1}{\eta}\log\ESum{\exp(\eta{\Psi}_{k}(x,a)))}{a}{A}. 
\end{align}

Finally, \textbf{(iii)} by plugging \eqref{PiPref} and \eqref{VPref}  into \eqref{PrefDef} we derive:
\begin{equation}
\label{MainODPPFirst}
\begin{aligned}
\Psi_{k+1}(x,a)={\Psi}_k(x,a)-\mathcal{L}_{\eta}\Psi_k(x)+r(x,a)+\gamma({P}\mathcal{L}_{\eta}\Psi_k)(x,a),
\end{aligned}
\end{equation}
with $\mathcal{L}_{\eta}$ operator  being defined by 
$\mathcal{L}_{\eta}\Psi(x)\triangleq1\big/\eta\log\sum_{\inset{a}{A}}{\exp(\eta{\Psi(x,a))}}$.
\eqref{MainODPPFirst} is one form of the DPP equations.  There is a more
efficient and analytically more tractable version of the DPP equation, where we
replace  $\mathcal{L}_{\eta}$ by the Boltzmann soft-max
$\mathcal{M}_{\eta}$ defined by
$\mathcal{M}_{\eta}\Psi(x)\triangleq\sum_{\inset{a}{A}}\big[{\exp(\eta{\Psi(x,a)})\Psi(x,a)}\big/\sum_{\inset{a'}{A}}{\exp(\eta{\Psi}(x,a'))}\big]$.\footnote{Replacing
$\mathcal{L}_\eta$ with $\mathcal{M}_{\eta}$ is motivated by the following
relation between these two operators:

\begin{align}
\label{LogSoftRel}
\begin{aligned}
|\mathcal{L}_\eta{\Psi}(x)-\M{\Psi}(x)|=1/{\eta}H_{\pi}(x)\leq\frac{L}{\eta}
\end{aligned}
,&&\forall\inset{x}{X},
\end{align}
with $H_{\pi}(x)$ is the entropy of the policy distribution $\pi$ obtained by plugging $\Psi$ into \eqref{PiPsiPref}. In words,  $\M{\Psi}(x)$ is  close to  $\mathcal{L}_\eta{\Psi}(x)$ up to the constant $L\big/\eta$. Also, both $\mathcal{L}_{\eta}\Psi(x)$ and $\M[\eta]\Psi(x)$ converge to $\M[]\Psi(x)$ when $\eta$ goes to $+\infty$. For the proof of~\eqref{LogSoftRel} and further readings see \citet[chap.~31]{MacKay03}. 
} In principle, we can provide formal analysis for
both versions.  However, the proof is somewhat simpler for the $\mathcal{M}_{\eta}$ case, which we will consider in the remainder of this paper.   By replacing $\mathcal{L}_{\eta}$ with $\mathcal{M}_{\eta}$ we deduce the  DPP recursion:
\begin{align}
\label{MainODPP}
\begin{aligned}
\Psi_{k+1}(x,a)=\mathcal{O}\Psi_k(x,a)&\triangleq{\Psi}_k(x,a)+r(x,a)+\gamma{P}\mathcal{M}_{\eta}\Psi_k(x,a)-\mathcal{M}_{\eta}\Psi_k(x)
\\&={\Psi}_k(x,a)+\T[\pi_k]\Psi_k(x,a)-\pi_k\Psi_k(x)
\end{aligned}
,&&\forall\inset{(x,a)}{Z},
\end{align}
where $\mathcal{O}$ is an operator defined on the action preferences $\Psi_k$ and $\pi_k$ is the soft-max policy  associated  with $\psi_k$:

\begin{equation}
\label{PiPsiPref}
\pi_k(a|x)\triangleq\frac{\exp(\eta{\Psi}_{k}(x,a))}{\ESum{\exp(\eta{\Psi}_{k}(x,a'))}{a'}{A}}.
\end{equation}

In Subsection~\ref{DPPProof}, we show that this iteration
gradually moves the policy towards the greedy optimal policy.         
Algorithm~\ref{DPPAlg1} shows the procedure. 

Finally, we would like to emphasize on an important difference between DPP and the double-loop algorithm resulted by solving~\eqref{ExpValKLMax1}.  One may notice that DPP algorithm, regardless of the choice of $\eta$,  is always incremental in $\psi$ and $\pi$  even when  $\eta$ goes to $+\infty$, whereas, in the case  of double-loop update of  the policy and the value function,  the algorithm is reduced to standard value iteration for $\eta=+\infty$ which is apparently not incremental in the policy $\pi$.  The reason for this difference is due to the extra smoothness introduced to DPP update rule by replacing the double-loop update with a single loop in~\eqref{ExpValIter} and~\eqref{ExpPolIter}.
\begin{algorithm}
\label{DPPAlg1}
\caption{ (DPP) Dynamic Policy Programming}
\KwIn{Randomized action preferences  $\Psi_0 (\cdot)$, $\gamma$ and $\eta$ }
\For{$k=0,1,2,\dots,K-1$}{
\For{each $\inset{ (x,a)}{Z}$}{
\For{each $\inset{a'}{A}$}{
$\pi_{k}(a'|x):=\frac{\displaystyle\exp(\eta{\Psi}_{k}(x,a'))}{\displaystyle\ESum{\exp(\eta{\Psi}_{k}(x,a''))}{a''}{A}}$;
}
$\Psi_{k+1}(x,a):=\Psi_k(x,a)+\T[\pi_k]\Psi_k(x,a)-\pi_k\Psi_k(x)$;
}
}
\For{each $\inset{ (x,a)}{Z}$}{
$\pi_{K}(a|x):=\frac{\displaystyle\exp(\eta{\Psi}_{K}(x,a))}{
\displaystyle\ESum{\exp(\eta{\Psi}_{K}(x,a'))}{a'}{A}}$;
}
\Return{$\pi_{K}$};
\end{algorithm} 
\subsection{Performance Guarantee}
\label{DPPProof}
We begin by proving a finite iteration performance guarantee for  DPP:

\begin{theorem}[ The $\ell_{\infty}$-norm performance loss bound of DPP]
\label{DPPMain1}
Let  Assumption~\ref{Regular} hold.  Also,  assume that $\Psi_0$ is uniformly bounded from above by $V_{\max}$ for all $\inset{(x,a)}{Z}$, then the following inequality holds for the policy induced by DPP at round $k\geq0$: 
\begin{equation*}
\Norm{Q^*-Q^{\pi_k}}\leq\frac{2\gamma\left(4V_{\max}+\frac{\log(L)}{\eta}\right)}{(1-\gamma)^2(k+1)}.
\end{equation*}
\end{theorem}
\begin{proof}
See Appendix~\ref{Thm1Prf}. 
\end{proof}
We can optimize this bound by the choice of $\eta=\infty$, for which the soft-max policy and the soft-max operator $\M$ are replaced with  the greedy policy and the max-operator $\M[]$.  As an immediate consequence of Theorem~\ref{DPPMain1}, we obtain the following result:
\begin{corollary}
\label{Q1}
The following relation holds in limit:
\begin{equation*}
\begin{aligned}
\subLim{k}{+\infty}{Q^{\pi_k}(x,a)}=Q^*(x,a),
&&
\forall\inset{(x,a)}{Z}. 
\end{aligned}
\end{equation*}
\end{corollary}
In words, the policy induced by DPP asymptotically converges to  the optimal policy $\pi^*$.   One can  also show that, under some mild assumption, there exists a unique limit for the action preferences in infinity. 
\begin{asm}
\label{uniq}
We assume that MDP has a unique deterministic optimal policy $\pi^*$ given by: 
\begin{equation*}
\begin{array}{cccc}
\pi^*(a|x)=
\left\{
\begin{array}{cccc}
\displaystyle
1&&&a=a^* (x)\\\displaystyle0&&&\mathrm{otherwise}
\end{array}
\right.,
&&&\forall\inset{x}{X},
\end{array}
\end{equation*}
where $a^* (x)=\arg\max_{ \inset{a}{A} }Q^*(x,a) $.  
\end{asm}
\begin{theorem}
 \label{DPP2}
Let  Assumption~\ref{Regular} and~\ref{uniq}  hold and   $k$  be a non-negative integer and let $\Psi_k(x,a)$, for all $\inset{(x,a)}{Z}$, be the action preference after $k$ iteration of DPP.  Then, we have:
\begin{equation*}
\begin{array}{cccc}
\subLim{k}{+\infty}\Psi_k(x,a)=\left\{
\begin{array}{cccc}
V^* (x)&a=a^* (x)
\\-\infty&\mathrm{otherwise}
\end{array}
\right.,
&
\forall{x}\in\mathcal{X}.
\end{array}
\end{equation*}
\end{theorem}
\begin{proof}
See Appendix~\ref{Thm2Prf}. 
\end{proof}
\section{Dynamic Policy Programming with Approximation}
\label{Approx}
Algorithm~\ref{DPPAlg1} (DPP) only applies to  small problems  with a  few  states and actions. One can   generalize  the DPP algorithm for the problems of practical scale by using function approximation techniques. Also, to compute the optimal policy by DPP  an explicit knowledge of model is required. In many real world problems,  this information is not available instead it may be possible to   simulate the state transition by Monte-Carlo sampling and then approximately \emph{estimate} the optimal policy using these samples.   In this section, we provide results on the performance-loss of DPP in the presence of approximation/estimation error.  We then compare $\ell_{\infty}$-norm performance-loss bounds of DPP with the standard results of AVI and API.  
Finally, We  introduce  new approximate algorithms for implementing DPP with Monte-Carlo sampling  (DPP-RL) and linear function approximation (SADPP).


\subsection{The $\ell_{\infty}$-norm performance-loss bounds for approximate DPP}
\label{ApproxBounds}

Let us consider a sequence of action preferences $\{\Psi_{0},{\Psi}_{1},{\Psi}_2,\dots\}$ such that, at round $k$, the action preferences  $\Psi_{k+1}$ is the result of approximately applying the DPP operator by the means of function approximation or Monte-Carlo simulation, i.e., for all $\inset{(x,a)}{Z}$: $\Psi_{k+1}(x,a)\approx\OP{\Psi_k(x,a)}$.   The  error term $\epsilon_k$ is defined as the difference of $\OP{\Psi_k}$  and its approximation:
\begin{align}
\label{DefineError}
\epsilon_k(x,a)\triangleq{\Psi}_{k+1}(x,a)-\OP{\Psi_k(x,a)},&&&&\forall\inset{(x,a)}{Z}.
\end{align}

The approximate DPP update rule is then given by :  
\begin{align}
\label{DPPApprox}
 \begin{aligned}
\OPMS[k+1]{\Psi}(x,a)&=\OPMS[k]{\Psi}(x,a)+{r}(x,a)+\gamma{P}{\mathcal{M}}_{\eta}\OPMS[k]{\Psi}(x,a)-{\mathcal{M}}_{\eta}\OPMS[k]{\Psi}(x,a)+{\epsilon}_{k}(x,a)
\\&=
\OPMS[k]{\Psi}(x,a)+\T[{\pi}_{k}]\OPMS[k]{\Psi}(x,a)-{\pi}_{k}\OPMS[k]{\Psi}(x,a)+{\epsilon}_{k}(x,a),
\end{aligned}
\end{align}
where $\pi_{k}$ is given by~\eqref{PiPsiPref}.

We begin by finite iteration analysis of the approximate DPP. The following theorem establishes an upper-bound on the performance loss of DPP in the presence of approximation error.  The proof is based on generalization of the bound that we established for DPP by taking into account the error $\epsilon_k$:
\begin{theorem}[$\ell_{\infty}$-norm performance loss bound of approximate DPP]
\label{ADPPMain1}

Let  Assumption~\ref{Regular} hold.  Assume that  $k$ is a non-negative integer and  $\Psi_0$  is bounded by $V_{\max}$. Further,   define $\epsilon_k$  for all k by~\eqref{DefineError} and the accumulated error  $E_k$ as:
\begin{equation}
\label{AveEpsDef}
E_k(x,a)\triangleq\sum_{j=0}^{k}\epsilon_j(x,a),\quad\quad\forall\inset{(x,a)}{Z}.
\end{equation}

Then the following inequality holds for the policy induced by approximate DPP at round $k$: 
\begin{equation*}
\Norm{Q^*-Q^{\pi_k}}\leq\frac{1}{(1-\gamma)(k+1)}\cl[[]{\frac{2\gamma\left(4V_{\max}+\frac{\log(L)}{\eta}\right)}{(1-\gamma)}+\sum_{j=0}^{k}\gamma^{k-j}\|E_{j}\|}.
\end{equation*}
\end{theorem}
\begin{proof}
See Appendix~\ref{Thm3Prf}. 
\end{proof}

Taking the  upper-limit yields in the following  corollary of Theorem~\ref{ADPPMain1}. 
\begin{corollary}[Asymptotic $\ell_{\infty}$-norm performance-loss bound of approximate DPP] 
\label{C2}
Define   $\bar{\varepsilon}\triangleq{\lim\sup}_{k\rightarrow\infty}\Norm{E_k}\big/(k+1)$. Then, the following inequality holds:
\begin{equation}
\label{AsymptADPP}
\underset{k\rightarrow\infty}{\lim\sup}\Norm{Q^*-Q^{\pi_k}}\leq\frac{2\gamma}{(1-\gamma)^2}\bar{\varepsilon}.
\end{equation}
\end{corollary}

The asymptotic bound is similar to the existing results of AVI and API~\citep[chap.  6]{Thiery10,Tsitsiklis96}:
\begin{equation*}
\underset{k\rightarrow\infty}{\lim\sup}\Norm{Q^*-Q^{\pi_k}}\leq\dfrac{2\gamma}{(1-\gamma)^2}\varepsilon_{\max},
\end{equation*}
where $\varepsilon_{\max}={\lim\sup}_{k\rightarrow\infty}\Norm{\epsilon_k}$.  The difference is that in~\eqref{AsymptADPP} the supremum norm of error  $\varepsilon_{\max}$ is replaced  by  the supremum norm of the average error $\bar{\varepsilon}$. In other words, unlike AVI and API, the  size of error at each iteration is not a critical factor for the performance of DPP and  as long as the size of  average error remains close to $0$, DPP can achieve a near-optimal performance even when the error itself is arbitrary large.  To gain a better understanding of this result consider a case  in which, for any algorithm, the sequence of  errors $\{\epsilon_0,\epsilon_1,\epsilon_2,\dots\}$   are  some i.i.d.  zero-mean random variables bounded by $0<U<\infty$. We then obtain the following asymptotic bound for  the approximate DPP by applying the law of large numbers to  Corollary~\ref{C2}:

\begin{equation}
\label{iidConv}
\begin{aligned}
\underset{k\rightarrow\infty}{\lim\sup}\Norm{Q^*-Q^{\pi_k}}&\leq\frac{2\gamma}{(1-\gamma)^2}\bar{\varepsilon}=0,\quad\quad\text{w.p. (with probability) 1},
\end{aligned}
\end{equation}
whilst for API and AVI we have:
\begin{equation*}
\begin{aligned}
\underset{k\rightarrow\infty}{\lim\sup}\Norm{Q^*-Q^{\pi_k}}&\leq\frac{2\gamma}{(1-\gamma)^2}U. 
\end{aligned}
\end{equation*}
In  words,  approximate DPP  manages to cancel the i.i.d. noise and asymptotically converges to the optimal policy whereas there is no guarantee,  in this case, for the convergence of  API and AVI to the optimal solution.  This result suggests  that DPP can  average out the simulation noise caused  by Monte-Carlo sampling and eventually achieve a significantly better performance  than AVI and API in the presence of large variance of estimation. We will show, in the the next subsection that a sampling-based  variant of DPP (DPP-RL) manages to cancel the simulation noise and asymptotically converges, almost surely, to the optimal policy (see Theorem~\ref{DPP_RLConv}). 

\subsection{Reinforcement Learning with Dynamic Policy Programming}
To compute the optimal policy by DPP one needs an explicit knowledge of model. In many problems we do not  have access to this information but instead we can generate samples by simulating  the model. The optimal policy can then be \emph{learned} using these samples. In this section, we introduce a new RL algorithm, called DPP-RL,  which relies on a sampling-based variant of  DPP to update the policy.  The update rule of DPP-RL is very similar to~\eqref{MainODPP}. The only difference is that,  in DPP-RL,  we replace the Bellman operator $\T^{\pi}\Psi(x,a)$  with its sample estimate  $\T_k^{\pi}\Psi(x,a)\triangleq{r}(x,a)+\pi\Psi(y_k)$, where the next sample $y_k$ is drawn from $P(\cdot|x,a)$:\footnote{We assume, hereafter, that we have access to the generative model of MDP, i.e., given the state-action pair $(x,a)$ we can generate the next sample $y$ from $P(\cdot|x,a)$ for all $\inset{(x,a)}{Z}$.}

\begin{equation}
\label{DPPRLup}
\Psi_{k+1}(x,a)\triangleq\Psi_k(x,a)+\T_k^{\pi_k}\Psi_k(x,a)-\pi_k\Psi_k(x).
\end{equation}

The pseudo-code of DPP-RL algorithm is shown in Algorithm~\ref{DPPRLAlg}.

\begin{algorithm}[H]
\label{DPPRLAlg}
\caption{(DPP-RL) Reinforcement learning with DPP }\label{alg:DPP}
\KwIn{ Initial the action preferences $\Psi_{0}(\cdot)$, discount factor $\gamma$,  $\eta$ and number of iterations $K$}
Generate a set of  i.i.d. samples $\{y_1,y_2,y_3,\dots,y_K\}$, for every $\inset{(x,a)}{Z}$, from ${P}(\cdot|x,a)$\;
\For{$k:=0,1,2,3,\dots,K-1$}{

\For{each $\inset{(x,a)}{Z}$}
{

\For{each $\inset{a'}{A}$}{
$\pi_{k}(a'|y_k)=\displaystyle{\dfrac{\exp(\eta\Psi_{k}(y_k,a'))}{\ESum{\exp(\eta\Psi_{k}(y_k,a''))}{a''}{A}}}$\;
}
$\T[\pi_k]_k{\Psi}_{k}(x,a):=r(x,a)+\gamma\pi_k{\Psi}_{k}(y_k)$\;
$\Psi_{k+1}(x,a):= \Psi_k(x,a) +\T[\pi_k]_k\Psi_k(x,a) - \pi_k\Psi_k(x)$\;
}
}
$\pi_K(a|x)=\dfrac{\exp(\eta\Psi_K(x,a))}{\ESum{\exp(\eta\Psi_K(x,a'))}{a'}{A}}$\;
\Return{$\pi_{K}$
}
\end{algorithm}

Equation~\eqref{DPPRLup} is just   an approximation of DPP update rule~\eqref{MainODPP}. Therefore, the convergence result of  Corollary~\ref{Q1} does not hold for DPP-RL. However, the new algorithm still  converges to the optimal policy since  one can show that the errors associated with approximating~\eqref{MainODPP}  are asymptotically \emph{averaged out} by DPP-RL, as postulated  by  Corollary~\ref{C2}.
The following theorem  establishes the  asymptotic convergence of the policy induced by DPP-RL to the optimal policy.

\begin{theorem}[Asymptotic convergence  of DPP-RL]
\label{DPP_RLConv}
Let Assumption~\ref{Regular} hold. Assume that the initial action-value function $\Psi_0$  is uniformly bounded by $V_{\max}$ and $\pi_k$ is the policy induced by Algorithm~\ref{DPPRLAlg} at round $k$. Then, w.p. 1, the following holds:
\begin{align*}
\lim_{k\rightarrow\infty}{Q^{\pi_k}}(x,a)=Q^*(x,a),\quad\quad\forall\inset{(x,a)}{Z}.
\end{align*}
\end{theorem}

\begin{proof}
See Appendix~\ref{DPP_RLProof}.
\end{proof}

One may notice  that the update rule of DPP-RL, unlike other incremental RL methods such as Q-learning~\citep{Watkins92} and SARSA~\citep{singh00}, does not involve any decaying learning step. This is an important difference since it is known that the convergence rate  of  incremental RL methods like Q-learning is very sensitive to the choice of learning step~\citep{Even-DarM03,Szepesvari97} and a bad choice of the learning step may lead to significantly  slow rate of convergence. DPP-RL  seems to not suffer from this problem (see Figure~\ref{fig:DPP-RL}) since the DPP-RL update rule is just an empirical estimate of the update rule of  DPP. Therefore, one may expect that the rate of convergence of DPP-RL  remains close to the fast rate of convergence of DPP established in Theorem~\ref{DPPMain1}.

\subsection{Approximate Dynamic Policy Programming with Linear Function Approximation}
In this subsection, we consider DPP with \emph{linear function approximation} (LFA)  and  \emph{least-squares regression}. Given a set of basis functions $\mathcal{F}_{\phi} = \cl[\{]{\phi_1, \dots, \phi_k}$, where each $\phi_i:\mathcal{Z}\to\Re$  is a bounded real  valued function, the sequence of action preferences $\{\Psi_{0},{\Psi}_{1},{\Psi}_2\cdots\}$  are defined as a linear combination of these basis functions: $\Psi_{k}=\theta^{\mathsf{T}}_{k}\Phi$,  where $\Phi$ is a ${m}\times{1}$ column vector with the entries $\{\phi_{i}\}_{i=1:m}$ and $\theta_k \in \Re^m$ is a $m \times1$ vector of  parameters. 

The action preference function $\Psi_{k+1}$ is an approximation of the DPP operator $\OP{\Psi}_{k}$. In case of LFA the  common approach to approximate DPP operator is to find a vector $\theta_{k+1}$ that projects  $\OP{\Psi_k}$ on the column space spanned by $\Phi$ by minimizing  the loss function:
\begin{equation}
\label{LSP}
J_k(\theta;\Psi)\triangleq\Norm[2,\mu]{{\theta}^\mathsf{T}\Phi-\OP{\Psi}_k}^{2},
\end{equation}
where $\mu$ is a probability measure on $\mathcal{Z}$.    The best solution, that minimize  $J$,  is  called  the  least-squares solution:
\begin{equation}
\label{LSSol}
\theta_{k+1}=\arg\subExt[min]{\theta}{\Re^m}J_k(\theta;\Psi)=\big[\Eval{\big(\Phi\Phi^{\mathsf{T}}\big)}\big]^{-1}\Eval{(\Phi\OP{\Psi}_k)},
\end{equation}
where the expectation is taken w.r.t. $(x,a)\sim\mu$. In principle, to compute the least squares solution equation requires to compute $\OP{\Psi}_k$ for all states and actions.  For large scale problems this becomes infeasible.  Instead, we can make a sample estimate of the least-squares solution by minimizing the empirical loss $\tilde{J}_k(\theta;\Psi)$:

\begin{equation*}
\label{LSPHat}
\tilde{J}_k(\theta;\Psi)\triangleq\frac{1}{N}\ESum[U]{(\theta^{\mathsf{T}}\Phi(X_n,A_n)-\OP_n{\Psi}_k)^{2}}{n=1}{N}+\alpha\theta^{\mathsf{T}}\theta,
\end{equation*}
where $\{(X_n,A_n)\}_{n=1:N}$ is a set of $N$ i.i.d. samples drawn  from the distribution $\mu$.  Also, $\OP_n{\Psi}_k$ denotes  a single sample estimate of $\OP{\Psi}_k(X_n,A_n)$  defined by $\OP_n{\Psi}_k\triangleq \Psi_k(X_n,A_n) + r(X_n,A_n) + \gamma\M[\eta]\Psi_k(X'_n) - \M[\eta]\Psi_k(X_n)$, where $X'_n \sim P(\cdot|X_n,A_n)$. Further, to avoid over-fitting due to the small size of data set, we add a quadratic regularization term to the loss function. The empirical least-squares solution which minimizes $\tilde{J}_k(\theta;\Psi)$ is given by:
\begin{equation}
\label{LSSample}
\tilde{\theta}_{k+1}=\cl[[]{\sum_{n=1}^N\Phi(X_n,A_n)\Phi(X_n,A_n)^{\mathsf{T}}+\alpha{N}\mathbf{I}}^{-1}\sum_{n=1}^N\OP_n \Psi_k\Phi(X_n,A_n).
\end{equation}

Algorithm~\ref{BatchADPPAlg} presents the \emph{sampling-based approximate dynamic policy programming} (SADPP) in which we rely  on~\eqref{LSSample} to approximate DPP operator at each iteration.

\begin{algorithm}[H]
\label{BatchADPPAlg}
\caption{ (SADPP) Sampling-based approximate dynamic policy programming}
\KwIn{$\tilde{\theta}_{0}$, $\eta$, $\gamma$, $\alpha$, $K$ and $N$}
\For{$k=0,1,2,\dots,K-1$}{
Generate a  set of  i.i.d. samples  $\{(X_n,A_n,X'_n)\}_{n=1:N}$ by drawing $N$ sample from $\mu$ and $P(\cdot|X_n,A_n)$\; 
\For{$n=1,2,3,\dots,N$}{
$\Psi_{k}(X_n,A_n)=\tilde\theta_{k}^{\mathsf{T}}\Phi(X_n,A_n)$\;
\For{each $\inset{A'}{A}$}{
$\Psi_{k}(X_n,A')=\tilde\theta_{k}^{\mathsf{T}}\Phi(X_n,A')$\;
$\Psi_{k}(X'_n,A')=\tilde\theta_{k}^{\mathsf{T}}\Phi(X'_n,A')$\;
}
$\M{\Psi}_{k}(X_n)=\ESum{\frac{\exp(\eta{\Psi}_k(X_n,A'))\Psi_k(X_n,A')}{\ESum{\exp\eta{\Psi}_k(X_n,A'')}{A''}{A}}}{A'}{A}$\;
$\M{\Psi}_{k}(X'_n)=\ESum{\frac{\exp(\eta{\Psi}_k(X'_n,A'))\Psi_k(X'_n,A')}{\ESum{\exp\eta{\Psi}_k(X'_n,A'')}{A''}{A}}}{A'}{A}$\;
$\OP_n{\Psi}_{k}={\Psi}_{k}(X_n,A_n)-r(X_n,A_n)-\gamma(\M{\Psi_{k}})(X'_n)+(\M{\Psi_{k}})(X_n)$\; 
}
$\tilde{\theta}_{k+1}=\cl[[]{\sum_{n=1}^N\Phi(X_n,A_n)\Phi(X_n,A_n)^{\mathsf{T}}+\alpha{N}\mathbf{I}}^{-1}\sum_{n=1}^N\OP_n{\Psi}_{k}\Phi(X_n,A_n)$\;
}
\Return{$\tilde{\theta}_{K}$}
\end{algorithm}

\section{Numerical Results}
\label{NumRes}
\input{numericalAug.tex}

\section{Related Work}
\label{rel}
There are other methods which rely on a incremental update of the policy.  One
well-known algorithm of this kind is the \emph{actor-critic} method (AC), in
which the actor uses the value function computed by the critic to guide the
policy search~\citep[chap.~6.6]{Sutton98}.  An important extension of AC, the
\emph{policy-gradient actor critic} (PGAC), extends the idea of AC to problems
of practical scale~\citep{Sutton00,Peters08}. In PGAC, the actor updates the
parameterized policy in the direction of the (natural) gradient of performance,
provided by the critic. The gradient update ensures that PGAC asymptotically
converges to a local maximum, given that an unbiased estimate of the gradient
is provided by the critic~\citep{MaeiSBS10,BhatnagarSGL09,Konda03,Kakade02}.  
Other incremental RL methods include Q-learning~\citep{Watkins92} and  SARSA~\citep{singh00} 
which can be considered as the incremental variants of the value iteration and the optimistic policy iteration algorithms, respectively~\citep{Tsitsiklis96}. These algorithms have been shown to converge to the optimal value function in tabular case~\citep{Tsitsiklis96,Jaakkola94}. 
Also, there are some studies in the literature concerning the asymptotic convergence of Q-learning  in the presence of  function approximation~\citep{Melo08,Szepesva04}. However, to the best of our knowledge, there is no preceding in the literature for asymptotic or  finite-iteration performance loss bounds of incremental RL methods  and  this study appears to be the first to prove such a bound for an incremental RL algorithm.

The work proposed in this paper has some relation to recent work by
\citet{Kappen05b} and \citet{Todorov06}, who formulate a stochastic optimal
control problem to find a conditional probability distribution $p (x'|x)$ given
an uncontrolled dynamics $\bar{p} (x'|x)$. The control cost is the relative
entropy between $p (x'|x)$ and $\bar{p} (x'|x)\exp (r(x))$. The difference is
that in their work a restricted class of control problems is considered for
which the optimal solution $p$ can be computed directly in terms of $\bar{p}$
without requiring Bellman-like iterations. Instead, the present approach is
more general, but does  requires Bellman-like iterations. Likewise, our
formalism is superficially similar to  PoWER~\citep{KoberP08} and
SAEM~\citep{VlassisT09}, which rely on EM algorithm to maximize a lower bound
for the expected return in an iterative fashion. This lower-bound also can be
written as a KL-divergence between two distributions. Another relevant study is
\emph{relative entropy policy search} (REPS)~\citep{PetersMA10} which relies on
the idea of minimizing the relative entropy to control the size of policy
update.   The main differences are: \textbf{(i)} the REPS algorithm is an
actor-critic type of algorithm, while DPP is more a policy iteration type of
method. \textbf{(ii)}  In
REPS the inverse temperature $\eta$ needs to be  optimized while DPP
converges to the optimal solution for any inverse temperature $\eta$, and
\textbf{(iii)} here we provide a convergence analysis of DPP, while there is no
convergence analysis in REPS.

\section{Discussion and Future Works}
\label{conclus}
We have presented a new approach,  dynamic policy programming
(DPP), to compute the  optimal policy in infinite-horizon discounted-reward MDPs.   We have theoretically proven
the convergence of DPP to the optimal policy for the
tabular case.  We have also provided  performance-loss
bounds for  DPP in the presence of approximation. The bounds have been  expressed in terms of supremum norm of average accumulated error as opposed to standard results for AVI and API which expressed in terms of supremum norm of the errors. We have then introduced  a new incremental model-free RL algorithm,
called DPP-RL, which relies on a sample estimate instance of DPP update rule to estimate the optimal policy.  We have proven the  asymptotic convergence of DPP-RL to the optimal policy and then have compared its, numerically, with the standard RL methods.  Experimental results on various  MDPs have been provided showing that, in all cases,  DPP-RL is superior to other  RL methods in terms of convergence rate. This may be due to the fact that DPP-RL, unlike other incremental RL methods,  does not rely on stochastic approximation for estimating the optimal policy and  therefore it does not suffer from the slow convergence caused by  the presence of the decaying learning step in stochastic approximation. 


In this work, we are only interested in the estimation of the optimal policy and not the problem 
of exploration. Therefore, we have not compared our algorithms to the PAC-MDP methods~\citep{StrehlLL09}, 
in which the choice of the exploration policy impacts the behavior of the learning algorithm. 
Also, in this paper, we have not compared our results
 with those of (PG)AC since they rely on a different kind of  sampling strategy:
Both DPP-RL and SADPP rely on a generative model for sampling, whereas AC 
makes use of some trajectories of the state-action pairs, generated by Monte-Carlo simulation, 
to estimate the optimal policy.

In this study, we  provide $\ell_{\infty}$-norm performance-loss bounds  for
approximate DPP.  However, most supervised learning and regression algorithms
rely on minimizing some form of $\ell_{p}$-norm error.  Therefore, it is natural
to search for a kind of performance bound that relies on the $\ell_{p}$-norm of
approximation error.  Following ~\citet{Remi05}, $\ell_{p}$-norm bounds for
approximate DPP can be established by providing a bound on the performance
loss of each component of value function under the policy induced by DPP.
This would be a topic for future research.  

Another  direction for future work  is to provide finite-sample \emph{probably approximately correct} (PAC) bounds for SADPP and DPP-RL in the spirit of previous theoretical results available for fitted value iteration  and fitted $Q$-iteration~\citep{Munos08,Antos07}. In the case of SADPP, this would require extending the error propagation result of Theorem~\ref{ADPPMain1} to an $\ell_2$-norm analysis and combining it with the standard regression bounds.

Finally, an important extension of our results  would be to apply DPP for  large-scale action  problems.   In that case, we need an efficient way to approximate $\M\Psi_k(x)$ in update rule \eqref{MainODPP} since computing the exact summations become expensive.   One idea is to sample estimate $\M{\Psi}_k(x)$ using  Monte-Carlo simulation~\citep[chap.~29]{MacKay03}, since $\M{\Psi}_k(x)$ is the expected value of $\Psi_k(x,a)$ under the soft-max policy $\pi_k$. 

\appendix
\section{Proof of Proposition~\ref{ConstLem}}
\label{AppLem1}

We first introduce the Lagrangian function $\mathcal{L}\cl{x;{\lambda_x}}:\mathcal{X}\to\Re$: 

\begin{equation*}
\mathcal{L}\cl{x;{\lambda_x}}=\ESum{\pol\left[{r}(x,a)+\gamma\cl{PV^{*}_{\bar{\pi}}}(x,a)\right]}{a}{A}-\dfrac{1}{\eta}\KL{\pol[T]}{\pol[T-]}
-{\lambda_{x}}\Biggl[\ESum{\pol}{a}{A}-1\Biggr].
\end{equation*}

The maximization in \eqref{ExpValKLMax1} can be expressed as maximizing the Lagrangian function $\mathcal{L}\cl{x,{\lambda_x}}$. The necessary condition for the extremum with respect to $\pol[T]$ is:

\begin{equation*}
0=\pdiff{\mathcal{L}\cl{x,\lambda_{x}}}{\pol}={r}(x,a)+\gamma\cl{PV^{*}_{\bar{\pi}}}(x,a)-\dfrac{1}{\eta}-\dfrac{1}{\eta}\log\cl{\frac{\pol}{\pol[-]}}-\lambda_{x},
\end{equation*}
which leads to:
\begin{align}
\label{polPrelim}
 \bar{\pi}^{*}(a|x)=\pol[-]\exp\cl{-\eta\lambda_{x}-1}\exp\cl[[]{\eta({r}(x,a)+\gamma\cl{PV^{*}_{\bar{\pi}}}(x,a))},
 &&
 \forall\inset{x}{X}.
\end{align}

The Lagrange multipliers can then be solved from the constraints:
\begin{equation*}
 1=\ESum{\bar{\pi}^{*}(a|x)}{a}{A}=\exp\cl{-\eta\lambda_{x}-1}\ESum{\pol[-]\exp\cl[[]{\eta({r}(x,a)+\gamma\cl{PV^{*}_{\bar{\pi}}}(x,a))}}{a}{A},
\end{equation*}
\begin{equation}
\label{lambdaFinal}
\lambda_{x}=\dfrac{1}{\eta}\log\ESum{
\pol[-]\exp\cl[[]{\eta({r}(x,a)+\gamma\cl{PV^{*}_{\bar{\pi}}}(x,a))}}{a}{A}-\dfrac{1}{\eta}.
\end{equation}

By plugging \eqref{lambdaFinal} into \eqref{polPrelim} we deduce:
\begin{align}
\label{polFinal}
\bar{\pi}^{*}(a|x)=\dfrac{\pol[-]
\exp\cl[[]{\eta({r}(x,a)+\gamma\cl{PV^{*}_{\bar{\pi}}}(x,a))}}{\ESum{\pol[-]
\exp\cl[[]{\eta({r}(x,a)+\gamma\cl{PV^{*}_{\bar{\pi}}}(x,a))}}{a}{A}},
&&
 \forall\inset{(x,a)}{Z}.
\end{align}

The results then follows by substituting \eqref{polFinal} in \eqref{ExpValKLMax1}.



\section{Proof of Theorem~\ref{DPPMain1}}
\label{Thm1Prf}

In this section, we provide a formal analysis of the convergence behavior of DPP. Our  objective is to establish a rate of convergence for the value function  of the policy  induced by DPP.  

Our main result is in the form of following finite-iteration performance-loss bound, for all $k\geq0$:
\begin{align}
\label{MainRes}
\|Q^*-{Q}^{\pi_k}\|\leq\frac{2\gamma\left(4V_{\max}+\frac{\log(L)}{\eta}\right)}{(1-\gamma)^2(k+1)}.
\end{align}

Here, $\qv[{\pi_k}]$ is the action-values under the   policy $\pi_k$ and  $\pi_k$ is the policy induced by DPP at step $k$. 

To derive \eqref{MainRes} one needs to relate ${Q}^{\pi_k}$  to the optimal $\qv$.  Unfortunately, finding a direct relation between  $\qv[{\pi}_k]$ and  $\qv$ is not an easy task. Instead, we relate $\qv[{\pi}_k]$ to ${Q}^*$ via an auxiliary action-value function ${Q}_k$, which we define below. In the remainder of this Section we take the following steps: \textbf{(i)} we  express  ${\Psi_k}$ in terms of  ${Q}_k$ in Lemma~\ref{PExp1}. \textbf{(ii)} we obtain an upper bound on the normed error  $\Norm{\qv-{Q}_k}$ in Lemma~\ref{DPSubLinEx}. Finally, \textbf{(iii)} we use these two  results to  derive a  bound on  the normed error $\Norm{\qv-\qv[{\pi}_k]}$. For the sake of readability, we skip the formal proofs of the Lemmas in this section since we prove a more general case in Section~\ref{Thm3Prf}. Further, in the sequel, we  repress the state(-action) dependencies  in our notation wherever these dependencies are clear, e.g., $\Psi(x,a)$ becomes ${\Psi}$, $Q(x,a)$ becomes $Q$.

Now let us define the auxiliary action-value function ${Q}_{k}$. The sequence of auxiliary action-value functions $\{Q_0,{{Q}_1},{Q}_2,\dots\}$ is obtained by iterating the  initial  $Q_0=\Psi_0$ from the following recursion:

\begin{equation}
\label{AuxDef1}
\qv[]_k=\frac{k-1}{k}\T^{\pi_{k-1}}\qv[]_{k-1}+\frac{1}{k}\T^{\pi_{k-1}}Q_0,
\end{equation}
where $\pi_{k}$ is the policy induced by the $k^{\text{th}}$ iterate of DPP.
 
Lemma~\ref{PExp1} relates $\Psi_k$ with ${Q}_k$:

\begin{lemma}
\label{PExp1}
Let $k$ be a positive integer.  Then, we have:

\begin{equation}
\label{PsiQ}
\begin{aligned}
{\Psi}_k&= kQ_{k}+Q_0-\pi_{k-1}((k-1)Q_{k-1}+Q_0).
\end{aligned}
\end{equation}
\end{lemma}

Now we focus on relating ${Q}_k$ and ${Q}^*$:

\begin{lemma}
\label{DPSubLinEx}
Let  Assumption~\ref{Regular} hold   and $L$ denotes the cardinality of $\mathcal{A}$  and   $k$ be  a positive integer,  also assume that  $\Norm{{\Psi}_0}\leq{V_{\max}}$  then the following inequality holds:

\begin{equation}
\label{QAuxDiff}
\left\|\qv-\qv[]_k\right\|\leq\frac{\gamma\cl{4V_{\max}+\frac{\log(L)}{\eta}}}{(1-\gamma)k}.
\end{equation}

\end{lemma}

Lemma~\ref{DPSubLinEx} provides an upper bound on the normed-error $\Norm{\qv[]_{k}-\qv}$. We make use of Lemma~\ref{DPSubLinEx}  to prove the main result of this Section:

\begin{equation*}
\begin{aligned}
\Norm{Q^*-Q^{{\pi}_k}}&=\Norm{Q^*-Q_{k+1}+Q_{k+1}-\T[\pi_k]Q^*+\T[\pi_k]Q^*-Q^{{\pi}_k}}
\\&\leq\Norm{Q^*-Q_{k+1}}+\Norm{Q_{k+1}-\T[\pi_k]Q^*}+\Norm{\T[\pi_k]Q^*-\T[\pi_k]Q^{{\pi}_k}}
\\&\leq\Norm{Q^*-Q_{k+1}}+\Norm{Q_{k+1}-\T[\pi_k]Q^*}+\gamma\Norm{Q^*-Q^{{\pi}_k}}.
\end{aligned}
\end{equation*}

By collecting terms we obtain:
\begin{equation*}
\label{DPPFinalEq}
\begin{aligned}
\Norm{Q^*-Q^{{\pi}_k}}&\leq\frac{1}{1-\gamma}\cl{\Norm{Q^*-Q_{k+1}}+\Norm{Q_{k+1}-\T[\pi_k]Q^*}}
\\&=\frac{1}{1-\gamma}\cl{\Norm{Q^*-Q_{k+1}}+\Norm{\frac{k}{k+1} \T[\pi_k]Q_k+\frac{1}{k+1}\T[\pi_k]Q_0-\T[\pi_k]Q^*}}
\\&\leq\frac{1}{1-\gamma}\cl{\Norm{Q^*-Q_{k+1}}+\frac{k}{k+1}\Norm{\T[\pi_k]Q^*-\T[\pi_k]Q_k}+\frac{1}{k+1}\Norm{\T[\pi_k]Q^*-\T[\pi_k]Q_0}}
\\&\leq\frac{1}{1-\gamma}\cl{\Norm{Q^*-Q_{k+1}}+\frac{\gamma{k}}{k+1}\Norm{Q^*-Q_k}+\frac{\gamma}{k+1}\Norm{Q^*-Q_0}}.
\end{aligned}
\end{equation*}

This combined with  Lemma~\ref{DPSubLinEx} completes the Proof.


\section{Proof of Theorem~\ref{DPP2} }
\label{Thm2Prf}
First, we note that  $\qv[]_k$ converges to $\qv$ (Lemma~\ref{DPSubLinEx}) and subsequently ${\pi}_k$  converges to ${\pi}^*$ by~\eqref{PolExpandEx}. Therefore, there exists a limit for $\OPMS[k]{\Psi}$ since $\OPMS[k]{\Psi}$ writes in terms of  $\qv[]_k$, $\qv[]_0$ and ${\pi}_{k-1}$ (Lemma~\ref{PExp1}).

Now, we compute the limit of $\OPMS[k]{\Psi}$.  $\qv[]_k$ converges to $\qv$ with a linear rate from Lemma~\ref{DPSubLinEx}. Also, we have ${V}^*={\pi}^*{Q}^*$ by definition of $V^*$ and $Q^*$. Then, by taking the  limit of~\eqref{PsiQ} we deduce:
\begin{align*}
\underset{k\rightarrow\infty}{\lim}\OPMS[k]{\Psi} (x,a)&=\underset{k\rightarrow\infty}{\lim}\cl[[]{k\qv (x,a)+Q_0(x,a)- (k-1){V}^*(x)-(\pi^*Q_0)(x)}
\\&=\underset{k\rightarrow\infty}{\lim}k(\qv (x,a)-{V}^*(x))
\\&\quad+Q_0(x,a)-(\pi^*Q_0)(x)+{V}^*(x).
\end{align*}

This combined with Assumption~\ref{uniq} completes the Proof. 
\section{Proof of Theorem~\ref{ADPPMain1}}
\label{Thm3Prf}
This Section provides a formal theoretical analysis of the performance of dynamic policy programming in the presence of approximation. 

Consider a sequence of the action preferences $\{\Psi_0,{\Psi}_{1},{\Psi}_{2},\dots\}$ as the iterates of~\eqref{DPPApprox}. Our  goal is to establish an $\ell_{\infty}$-norm performance loss bound  of the policy  induced by approximate DPP. 
The main result  is that at  iteration $k\geq{0}$ of approximate DPP, we have:

\begin{equation}
\label{MainResApp}
\left\|\qv-\qv[\pi_k]\right\|\leq\frac{2}{(1-\gamma)(k+1)}\cl[[]{\frac{\gamma\left(4V_{\max}+\frac{\log(L)}{\eta}\right)}{(1-\gamma)}+\sum_{j=1}^{k+1}\gamma^{k-j+1}\|E_{j-1}\|},
\end{equation}
where  $E_{k}= \sum_{j=0}^{k}{\epsilon_k}$ is the cumulative approximation error up to step $k$. Here, $\qv[{\pi_k}]$ denotes the action-value function of  the   policy $\pi_k$ and  $\pi_k$ is the soft-max policy associated with $\Psi_{k}$.

As in the proof of Theorem~\ref{DPPMain1}, we relate $\qv$ with $\qv[{\pi}_k]$ via an auxiliary action-value function ${Q}_k$. In the sequel, we first express  $\Psi_k$ in terms of  $Q_k$ in Lemma~\ref{PExpApp}. Then, we obtain an upper bound on the normed error  $\Norm{\qv-{Q}_k}$ in Lemma~\ref{DPSubLinApp}. Finally, we use these two  results to  derive~\eqref{MainResApp}. 

Now, let us define the auxiliary action-value function ${Q}_{k}$. The sequence of auxiliary action-value function $\{{Q}_0,{Q}_1,{Q}_2,\dots\}$ is resulted by iterating the initial action-value function $Q_0=\Psi_0$ from the following recursion:
\begin{equation}
\label{AuxDefApp}
\begin{aligned}
{Q}_{k}=\frac{k-1}{k}\T^{\pi_{k-1}}\qv[]_{k-1}+\frac{1}{k}(\T^{\pi_{k-1}}Q_0+E_{k-1}),
\end{aligned}
\end{equation}
where~\eqref{AuxDefApp} may be considered as an approximate version of~\eqref{AuxDef1}. Lemma~\ref{PExpApp} relates $\Psi_k$ with ${Q}_k$:

\begin{lemma}
\label{PExpApp}
Let $k$ be a positive integer and $\pi_k$ denotes the policy induced by the approximate DPP at iteration $k$.   Then we have:
\begin{equation}
\label{OPQAA}
\begin{aligned}
{\Psi}_k&= k\qv[]_{k}+Q_0-{\pi}_{k-1}\left((k-1)\qv[]_{k-1}+Q_0\right).
\end{aligned}
\end{equation}
\end{lemma}
 
\begin{proof} 
We rely on induction for the  Proof of this Theorem. The result  holds for $k=1$ since one can easily show that~\eqref{OPQAA} reduces to~\eqref{DPPApprox}. We then show that if~\eqref{OPQAA} holds for $k$ then it also holds for $k+1$. From~\eqref{DPPApprox} we have:

\begin{align*}
\Psi_{k+1}&=\Psi_{k}+\T[\pi_k]{\Psi}_k-\pi_k{\Psi}_k+\epsilon_{k}
\\&= kQ_{k}+Q_0-\pi_{k-1}((k-1)Q_{k-1}+Q_0)+\T[\pi_k](kQ_{k}+Q_0-\pi_{k-1}((k-1)Q_{k-1}+Q_0))
\\&\quad-\pi_k(kQ_{k}+Q_0-\pi_{k-1}((k-1)Q_{k-1}+Q_0))+\epsilon_k
\\&= kQ_{k}+Q_0+\T[\pi_k](kQ_{k}+Q_0-\pi_{k-1}((k-1)Q_{k-1}+Q_0))-\pi_k(kQ_{k}+Q_0)
\\&\quad+E_k-E_{k-1},
\end{align*}
where in the last step we make use of the following:
\begin{equation*}
\pi_k\pi_{k-1}(\cdot)=\pi_{k-1}(\cdot),\quad\quad\T[\pi_k]\pi_{k-1}(\cdot)=\T[\pi_{k-1}](\cdot).
\end{equation*}

By collecting terms we deduce:
\begin{align*}
\Psi_{k+1}&=kQ_{k}-(k-1)\T[\pi_{k-1}]Q_{k-1}-\T[\pi_{k-1}]Q_0-E_{k-1}+k\T[\pi_k]Q_{k}+\T[\pi_k]Q_0+E_k
\\&\quad+Q_0-\pi_k(kQ_{k}+Q_0)
\\&=(k+1)Q_{k+1}+Q_0-\pi_k(kQ_k+Q_0),&&\text{by~\eqref{AuxDefApp}}.
\end{align*}

Thus~\eqref{OPQAA} holds for $k+1$, and is thus true for all $k\geq{1}$. 
\end{proof}

Based on Lemma~\ref{PExpApp}, one can express the policy induced by DPP, $\pi_{k}$, in terms of  $Q$:

\begin{equation}
\begin{aligned}
\label{PolExpandEx}
\pi_k(a|x)&=\frac{\exp\cl{\eta\cl{k{Q}_k(x,a)+{Q}_0(x,a)-\pi_{k-1}((k-1)Q_{k-1}+Q_0)(x)}}}{Z (x)}
\\&=\frac{\exp\cl{\eta\cl{{k{Q}_k(x,a)+{Q}_0(x,a)}}}}{Z' (x)},
\end{aligned}
\end{equation}
where $Z' (x)=Z (x)\exp\cl{\eta\pi_{k-1}((k-1)Q_{k-1}+Q_0)(x)}$ is the  normalization factor.  Equation \eqref{PolExpandEx} expresses ${\pi}_k$ in terms of  $Q_k$ and $Q_0$. 
In an analogy to Lemma~\ref{DPSubLinEx} we state the following lemma that  establishes a bound on $\Norm{{Q}^*-Q_k}$ 


\begin{lemma}[ $\ell_{\infty}$-norm bound on ${Q}^*-{Q}_k$]
\label{DPSubLinApp}
Let  Assumption~\ref{Regular} hold. Define $\qv[]_k$ by~\eqref{AuxDefApp}. Let  $L$ denotes the cardinality of $\mathcal{A}$  and   $k$ be  a non-negative integer,  also, assume that $\Norm{{\Psi}_0}\leq{V}_{\max}$, then the following inequality holds:
\begin{equation*}
\left\|\qv-\qv[]_k\right\|\leq\frac{\gamma\left(4V_{\max}+\frac{\log(L)}{\eta}\right)}{(1-\gamma)k}+\frac{1}{k}\sum_{j=1}^k\gamma^{k-j}\|E_{j-1}\|.
\end{equation*}
\end{lemma}

We make use of the following results to prove Lemma~\ref{DPSubLinApp}.

\begin{lemma} 
\label{SoftMaxDiv}
Let ${\eta}>0$ and $\mathcal{Y}$ be a finite set with cardinality $L$. Also assume that $\mathcal{F}$ denotes the space of measurable functions on $\mathcal{Y}$ with $\mathcal{Y}^f_{\max}\subset{\mathcal{Y}}$ set of all entries of $\mathcal{Y}$ which maximize $f$.  Then the following inequality holds for all $\inset{f}{F}$:
\begin{equation*}
\max_{\inset{y}{Y}}f(y)-\ESum{\frac{\exp({\eta}{f(y)})f(y)}{\ESum{\exp({\eta}{f(y')})}{y'}{Y}}}{y}{Y}\leq\frac{\log(L)}{{\eta}}.
\end{equation*}
\end{lemma}  

\begin{lemma}
\label{SoftBellmanDiff}
Let $\eta>0$ and $k$ be a positive integer. Assume $\Norm{Q_0}\leq{V_{\max}}$, then the following holds:

\begin{equation*}
\left\|k\T{Q}_k+\T{Q}_0-k\T[\pi_k]Q_k-\T[\pi_k]Q_{0}\right\|\leq\gamma\cl{2V_{\max}+\frac{\log(L)}{\eta}}.
\end{equation*}

\end{lemma}

\begin{proof}
(Proof of Lemma~\ref{DPSubLinApp})
We rely on induction for the proof of this Lemma. Obviously the result holds for $k=0$. Then we need to show that if~\eqref{QAuxDiff} holds for $k$ it also holds for $k+1$:

\begin{equation}
\label{QAuxApp}
\begin{aligned}
\left\|Q^*-Q_{k+1}\right\|&=\left\|\T{Q}^*-\left(\frac{k}{k+1}\T[\pi_k]Q_{k}+\frac{1}{k+1}(\T[\pi_{k}]Q_0+E_k)\right)\right\|
\\&=\left\|\frac{1}{k+1}(\T{Q}^*-\T[\pi_{k}]Q_0)+\frac{k}{k+1}(\T{Q}^*-\T[\pi_k]Q_{k})-\frac{1}{k+1}E_k\right\|
\\&=\frac{1}{k+1}\left\|\T{Q}^*-\T{Q_0}+\T{Q_0}-\T[\pi_{k}]Q_0+k(\T{Q}^*-\T{Q_k}+\T{Q_k}-\T[\pi_k]Q_{k})\right\|
\\&\quad+\frac{1}{k+1}\left\|E_k\right\|
\\&\leq\frac{1}{k+1}\cl[[]{\left\|\T{Q}^*-\T{Q}_0\right\|+\left\|k\T{Q}_k+\T{Q}_0-k\T[\pi_k]Q_k-\T[\pi_k]Q_{0}\right\|}
\\&\quad+\frac{k}{k+1}\left\|\T{Q}^*-\T{Q}_{k}\right\|+\frac{1}{k+1}\left\|E_k\right\|
\\&\leq\frac{1}{k+1}\cl[[]{\gamma\left\|{Q}^*-{Q}_0\right\|+\left\|k\T{Q}_k+\T{Q}_0-k\T[\pi_k]Q_k-\T[\pi_k]Q_{0}\right\|}
\\&\quad+\frac{\gamma{k}}{k+1}\left\|{Q}^*-{Q}_{k}\right\|+\frac{1}{k+1}\left\|E_k\right\|,
\end{aligned}
\end{equation}
Now based on Lemma~\ref{SoftBellmanDiff} and by plugging~\eqref{QAuxDiff} into~\eqref{QAuxApp} we have:
\begin{equation*}
\begin{aligned}
\left\|Q^*-Q_{k+1}\right\|&\leq\frac{\gamma}{k+1}\cl[[]{{4V_{\max}+\frac{\log{L}}{\eta}}}+\frac{\gamma{k}}{k+1}\cl[[]{\frac{\gamma\cl{4V_{\max}+\frac{\log(L)}{\eta}}}{k(1-\gamma)}+\frac{1}{k}\sum_{j=1}^k\gamma^{k-j}\|E_{j-1}\|}
\\&\quad+\frac{1}{k+1}\|E_k\|
\\&=\frac{\gamma\cl{4V_{\max}+\frac{\log(L)}{\eta}}}{(1-\gamma)(k+1)}+\frac{1}{k+1}\sum_{j=1}^{k+1}\gamma^{k-j+1}\|E_{j-1}\|.
\end{aligned}
\end{equation*} 

The result then follows, for all $k\geq{0}$, by induction.
\end{proof}

\begin{proof}(Proof of Lemma~\ref{SoftMaxDiv})
For any $\inset{f}{F}$ we have:
\begin{equation*}
\max_{\inset{y}{Y}}f(y)-\ESum{\frac{\exp({\eta}{f(y)})f(y)}{\ESum{\exp({\eta}{f(y')})}{y'}{Y}}}{y}{Y}=\ESum{\frac{\exp(-{\eta}{g(y)})g(y)}{\ESum{\exp(-{\eta}{g(y')})}{y'}{Y}}}{y}{Y},
\end{equation*}
with ${g}(y)=\max_{\inset{y}{Y}}f(y)-f(y)$. According to~\citet[chap.~31]{MacKay03}:

\begin{equation*}
\ESum{\frac{\exp(-{\eta}{g(y)})g(y)}{\ESum{\exp(-{\eta}{g(y')})}{y'}{Y}}}{y}{Y}=-\frac{1}{{\eta}}\log\ESum{\exp(-{\eta}{g}(y))}{y}{Y}+\frac{1}{{\eta}}H_{p},
\end{equation*}
where $H_p$ is the entropy of probability distribution $p$ defined by:
\begin{equation*}
p(y)=\dfrac{\exp(-{\eta}{g}(y))}{\ESum{\exp(-{\eta}{g}(y')}{y'}{Y})}.
\end{equation*}

The following steps complete the proof.
\begin{equation*}
\begin{aligned}
\ESum{\frac{\exp(-{\eta}{g(y)})g(y)}{\ESum{\exp(-{\eta}{g(y')})}{y'}{Y}}}{y}{Y}&\leq-\frac{1}{{\eta}}\log\cl[[]{1+\sum_{y\notin\mathcal{Y}^{f}_{\max}}\exp(-{\eta}{g}(y)))}+\frac{1}{{\eta}}H_{p}
\\&\leq
\frac{1}{{\eta}}H_{p}\leq\frac{\log(L)}{{\eta}}.
\end{aligned}
\end{equation*}
\end{proof}

\begin{proof}(Proof of Lemma~\ref{SoftBellmanDiff})
We have, by definition of operator $\T$:
\begin{equation}
\label{SoftBellmanDiffEq}
\begin{aligned}
\left\|k\T{Q}_k+\T{Q}_0-k\T[\pi_k]Q_k-\T[\pi_k]Q_{0}\right\|&\leq\gamma\left\|kP\M[]{Q}_k+P\M[]{Q}_0-kP^{\pi_k}Q_k-P^{\pi_k}Q_{0}\right\|
\\&=\gamma\left\|P(\M[]k{Q}_k+\M[]{Q}_0-{\pi_k}(kQ_k+Q_{0}))\right\|
\\&\leq\gamma\left\|\M[]k{Q}_k+\M[]{Q}_0-{\pi_k}(kQ_k+Q_{0})\right\|
\\&\leq\gamma\left\|2\M[]{Q}_0+\M[](k{Q}_k+{Q}_0)-{\pi_k}(kQ_k+Q_{0})\right\|
\\&\leq\gamma\cl{2\left\|Q_0\right\|+\left\|\M[](k{Q}_k+{Q}_0)-\M(kQ_k+Q_{0})\right\|},
\end{aligned}
\end{equation}
where in the last line we make use of  Equation~\eqref{PolExpandEx}. The result then follows by comparing~\eqref{SoftBellmanDiffEq} with Lemma~\ref{SoftMaxDiv}.

\end{proof}

Lemma~\ref{DPSubLinApp} provides an upper-bound on the normed-error $\Norm{\qv-\qv[]_{k}}$. We make use of this result to derive a  bound on the performance loss $\Norm{\qv-\qv[{{\pi}_k}]}$:

\begin{equation*}
\begin{aligned}
\Norm{Q^*-Q^{{\pi}_k}}&=\Norm{Q^*-Q_{k+1}+Q_{k+1}-\T[\pi_k]Q^*+\T[\pi_k]Q^*-Q^{{\pi}_k}}
\\&\leq\Norm{Q^*-Q_{k+1}}+\Norm{Q_{k+1}-\T[\pi_k]Q^*}+\Norm{\T[\pi_k]Q^*-\T[\pi_k]Q^{{\pi}_k}}
\\&\leq\Norm{Q^*-Q_{k+1}}+\Norm{Q_{k+1}-\T[\pi_k]Q^*}+\gamma\Norm{Q^*-Q^{{\pi}_k}}.
\end{aligned}
\end{equation*}

By collecting terms we obtain:
\begin{equation*}
\label{DPPFinalEq}
\begin{aligned}
\Norm{Q^*-Q^{{\pi}_k}}&\leq\frac{1}{1-\gamma}\cl{\Norm{Q^*-Q_{k+1}}+\Norm{Q_{k+1}-\T[\pi_k]Q^*}}
\\&=\frac{1}{1-\gamma}\cl{\Norm{Q^*-Q_{k+1}}+\Norm{\frac{k}{k+1} \T[\pi_k]Q_k+\frac{1}{k+1}(\T[\pi_k]Q_0+E_k)-\T[\pi_k]Q^*}}
\\&\leq\frac{1}{1-\gamma}\cl{\Norm{Q^*-Q_{k+1}}+\frac{k}{k+1}\Norm{\T[\pi_k]Q^*-\T[\pi_k]Q_k}+\frac{1}{k+1}\Norm{\T[\pi_k]Q^*-\T[\pi_k]Q_0}}
\\&\quad+\frac{1}{(1-\gamma)(k+1)}\|E_k\|
\\&\leq\frac{1}{1-\gamma}\cl{\Norm{Q^*-Q_{k+1}}+\frac{\gamma{k}}{k+1}\Norm{Q^*-Q_k}+\frac{1}{k+1}\|E_k\|+\frac{\gamma}{k+1}\Norm{Q^*-Q_0}}.\end{aligned}
\end{equation*}

This combined with  Lemma~\ref{DPSubLinApp} completes the proof.

\section{Proof of Theorem~\ref{DPP_RLConv}}
\label{DPP_RLProof}
We begin the analysis by introducing some new notations. We define the estimation error associated with the $k^{\mathrm{th}}$ iterate of DPP-RL as the difference between the Bellman operator $\T[\pi_k]\Psi_k(x,a)$ and its sample estimate:

\begin{equation*}
\epsilon_k(x,a)=\T[\pi_k]_{k}\Psi_k(x,a)-\T[\pi_k]\Psi_k(x,a),\quad\forall\inset{(x,a)}{Z}.
\end{equation*}

The DPP-RL update rule can then be re-expressed in  form of the more general approximate DPP update rule:
\begin{align*}
\label{DPPEstApprox}
 \begin{aligned}
\OPMS[k+1]{\Psi}(x,a)&=
\OPMS[k]{\Psi}(x,a)+\T[{\pi}_{k}]\OPMS[k]{\Psi}(x,a)-{\pi}_{k}\OPMS[k]{\Psi}(x,a)+{\epsilon}_{k}(x,a).
\end{aligned}
\end{align*}

Now let us define $\mathcal{F}_k$ as the filtration generated by the sequence of all random variables $\{y_1,y_2,y_3,\dots,{y}_k\}$ drawn from the distribution $P(\cdot|x,a)$ for all $\inset{(x,a)}{Z}$. We have the property that $\Eval({\epsilon_k(x,a)|\mathcal{F}_{k-1}})=0$
which means that for all $\inset{(x,a)}{Z}$ the sequence of estimation errors $\{\epsilon_1,\epsilon_2,\dots,\epsilon_k\}$ is a martingale difference sequence w.r.t. the filtration $\mathcal{F}_k$.  The asymptotic converge of DPP-RL to the optimal policy follows by extending the result of~\eqref{iidConv} to the case of bounded martingale differences. For that we need to show that the sequence of  estimation errors $\{\epsilon_j\}_{j=0:k}$ is uniformly bounded:

\begin{lemma}[Stability of DPP-RL]
\label{stability}
Let Assumption~\ref{Regular} hold and assume that the initial action-preference function $\Psi_0$  is uniformly bounded by $V_{\max}$, then  we have, for all $k\geq 0$,
\begin{align*}
\Norm{\T[\pi_k]_k{\Psi}_k}&\leq\frac{2\gamma\log{L}}{\eta(1-\gamma)}+V_{\max}, \quad\quad
\Norm{\epsilon_k}\leq\frac{4\gamma\log{L}}{\eta(1-\gamma)}+{2}{V}_{\max}.
\end{align*}
\end{lemma}

\begin{proof}
We first prove that $\|\T_k^{\pi_k}\Psi_k\|\leq{V}_{\max}$ by induction. Let us assume that the bound $\|\T_k\Psi_k\|\leq{V}_{\max}$ holds. Thus
\begin{align*}
\|\T[\pi_k]_{k+1}\Psi_{k+1}\|&\leq\Norm{r}+\gamma\Norm{P^{\pi_k}\Psi_{k+1}}\leq\Norm{r}+\gamma\Norm{\M\Psi_{k+1}}
\\&=\Norm{r}+\gamma\Norm{\M\cl{\Psi_k+\T[\pi_k]_k\Psi_{k}-\M\Psi_k}}\\&\leq\Norm{r}+\gamma\Norm{\M\cl{\Psi_k+\T[\pi_k]_k\Psi_{k}-\M\Psi_k}-\M[]\cl{\Psi_k+\T[\pi_k]_k\Psi_{k}-\M\Psi_k}}
\\&\quad+\gamma\Norm{\M[]\cl{\Psi_k+\T[\pi_k]_k\Psi_{k}-\M\Psi_k}}
\\&\leq\Norm{r}+\frac{\gamma\log{L}}{\eta}+\gamma\Norm{\M[]\cl{\Psi_k+\T[\pi_k]_k\Psi_{k}-\M\Psi_k}}
\\&=\Norm{r}+\frac{\gamma\log{L}}{\eta}+\gamma\Norm{\M[]\cl{\Psi_k+\T[\pi_k]_k\Psi_{k}-\M\Psi_k+\M[]\Psi_k-\M[]\Psi_k}}
\\&\leq\Norm{r}+\frac{\gamma\log{L}}{\eta}+\gamma\Norm{\M[](\M[]\Psi_k-\M\Psi_k)}+\gamma\Norm{\M[]\cl{\Psi_k-\M[]\Psi_k}}
\\&\quad+\gamma\Norm{\M[]\T[\pi_k]_k\Psi_{k}}
\\&\leq\Norm{r}+\frac{2\gamma\log{L}}{\eta}+\gamma\Norm{\T[\pi_k]_k\Psi_{k}}\leq\|r\|+\frac{2\gamma\log{L}}{\eta}+\frac{2\gamma^2\log(L)}{\eta(1-\gamma)}+\gamma{V}_{\max}
\\&\leq\frac{2\gamma\log{L}}{\eta(1-\gamma)}+R_{\max}+\gamma{V}_{\max}=\frac{2\gamma\log{L}}{\eta(1-\gamma)}+V_{\max},
\end{align*}
where we make use of   Lemma~\ref{SoftMaxDiv} to bound the difference between the max operator $\M[](\cdot)$ and the soft-max operator $\M(\cdot)$. Now, by induction, we deduce that  for all $k\geq 0$, $\|\T_{k}\Psi_{k}\|\leq2\gamma\log{L}\big/(\eta(1-\gamma))+V_{\max}$. The bound on $\epsilon_k$ is an immediate consequence of this result.
\end{proof}

Now based on Lemma~\ref{stability} and Corollary~\ref{C2} we prove the main result. We begin by recalling the result of  Corollary~\ref{C2}:

\begin{equation*}
\underset{k\rightarrow\infty}{\lim\sup}\Norm{Q^*-Q^{\pi_k}}\leq\dfrac{2\gamma}{(1-\gamma)^2}\lim_{k\rightarrow\infty}\frac{1}{k+1}\Norm{E_k},
\end{equation*}

Thus to prove the convergence of DPP-RL we only need to show that $1/(k+1)\Norm{E_k}$ asymptotically converges to $0$ w.p. 1. For this we rely on the strong law of large numbers for martingale differences~\citep{Hoffman76}, which states that the average of a sequence of martingale differences asymptotically converges, almost surely, to $0$  if  the second moments of all entries of the sequence are bounded by some  $0\leq{U}\leq\infty$. This is the case  for the  sequence of martingales $\{\epsilon_1,\epsilon_2,\dots\}$ since we already have proven the boundedness of $\|\epsilon_k\|$ in Lemma~\ref{stability}.  Thus, we deduce:

\begin{equation*}
\lim_{k\rightarrow\infty}\frac{1}{k+1}|E_k(x,a)|=0,\quad\quad \text{ w.p. 1}. 
\end{equation*}

Thus:
\begin{equation}
\label{ErrConv}
\lim_{k\rightarrow\infty}\frac{1}{k+1}\Norm{E_k}=0,\quad\quad \text{ w.p. 1}. 
\end{equation}

The result then follows by combining~\eqref{ErrConv}  with Corollary~\ref{C2}.

\vskip 0.2in
\bibliography{Refs}

\end{document}

%% file: numericalAug.tex
In this section, we analyze empirically the effectiveness of the proposed
algorithms on different problem domains.  We first examine the convergence
properties of DPP-RL (Algorithm~\ref{DPPRLAlg}) on several discrete
state-action problems and compare it with two standard algorithms: a
synchronous variant of Q-learning~\citep{Even-DarM03} (QL) and the model-based
\emph{Q}-value iteration (VI) of ~\citet{Kearns99}.  Next, we investigate the
finite-time performance of SADPP (Algorithm~\ref{BatchADPPAlg}) in the presence
of function approximation and a limited sampling budget per iteration.  In this
case, we consider a variant of the optimal replacement problem described
in~\citet{Munos08} and compare our method with  regularized   least-squares
fitted $Q$-iteration (RFQI) \citep{FarahmandGSM08}. The source code of all tested algorithms are freely available in~\texttt{http://www.mbfys.ru.nl/\textasciitilde{mazar}/{Research\textunderscore{Top}.html}}.

\subsection{DPP-RL}
\label{NumResDPPRL}
We consider the following large-scale MDPs as benchmark problems:
\begin{figure}
\begin{center}
\includegraphics[width=.55\textwidth]{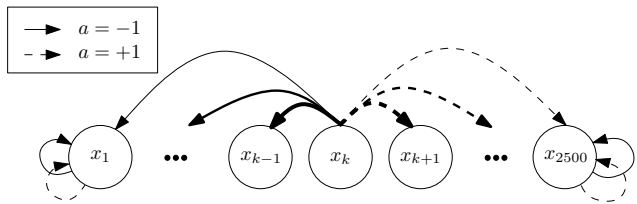}
\end{center}
\caption{
Linear MDP: Illustration of the linear MDP problem.  Nodes indicate states.
States $x_1$ and $x_{2500}$ are the two absorbing states and state $x_k$ is an
example of interior state.  Arrows indicate possible transitions of
\textbf{these three nodes only}.  From $x_k$ any other node is reachable with
transition probability (arrow thickness) proportional to the inverse of the
distance to $x_k$ (see the text for details).
}
\label{fig:linear}
\end{figure}
\begin{description}
\item[Linear MDP:]
this problem consists of states $\inset{x_k}{X}, k=\{1,2,\dots,2500\}$ arranged in a
one-dimensional chain (see Figure \ref{fig:linear}).  There are two possible
actions $\mathcal{A}=\{-1,+1\}$ (left/right) and every state is accessible from
any other state except for the two ends of the chain, which are absorbing
states. A state $\inset{x_k}{X}$ is called absorbing if $P(x_k|x_k,a)=1$ for
all $\inset{a}{A}$ and $P(x_l|x_k,a)=0, \forall l\neq k$.  Any transition to
one of these two states has associated reward $1$.

The transition probability for an interior state $x_k$ to any other state $x_l$
is inversely proportional to their distance in the direction of the selected
action, and zero for all states corresponding to the opposite direction.
Formally, consider the following quantity $n(x_l,a,x_k)$ assigned to all
non-absorbing states $x_k$ and to every $\inset{(x_l,a)}{Z}$:
\begin{equation*}
n(x_l,a,x_k)=
\begin{cases}
\dfrac{1}{|l-k|} & \text{for $(l-k)a>0$}\\
0                & \text{otherwise}
\end{cases}.
\end{equation*}
We can write the transition probabilities as:
\begin{equation*}
P(x_l|x_k,a)=
\dfrac{n(x_l,a,x_k)}{\ESum{n(x_m,a,x_k)}{x_m}{X}}.
\end{equation*}
Any transition that ends up in one of the interior states has associated reward
$-1$.

The optimal policy corresponding to this problem is to reach the closest
absorbing state as soon as possible.


\item[Combination lock:]
the combination lock problem considered here is a stochastic variant of the
reset state space models introduced in \cite{Koenig93}, where more than one
reset state is possible (see Figure \ref{fig:combclock}).  

In our case we consider, as before,  a set of states $\inset{x_k}{X},
k\in\{1,2,\dots,2500\}$ arranged in a one-dimensional chain and two possible actions
$\mathcal{A}=\{-1,+1\}$.  In this problem, however, there is only one absorbing
state (corresponding to the state \emph{lock-opened}) with associated reward of $1$.
This state is reached if the all-ones sequence $\{+1,+1,\dots,+1\}$ is entered
correctly.  Otherwise, if at some state $x_k$, $k<2500$, action $-1$ is taken,
the lock automatically resets to some previous state $x_l$, $l<k$ randomly (in
the original combination lock problem, the reset state is always the initial
state $x_1$).

For every intermediate state, the rewards of actions $-1$ and $+1$ are set to
$0$ and $-0.01$, respectively.  The transition probability upon taking
the wrong action $-1$ is, as before, inversely proportional to the
distance of the states. That is
\begin{align*}
n(x_k,x_l)&=
\begin{cases}
\dfrac{1}{k-l} & 
\text{for $l<k$}\\
0 &\text{otherwise}
\end{cases},
&
P(x_l|x_k,0)&=
\dfrac{n(k,l)}{\ESum{n(k,m)}{x_m}{X}}.
\end{align*} 
%
%
%

%
%
%
Note that this problem is more difficult than the linear MDP 
since the goal state is only reachable
from one state, $x_{2499}$.
\begin{figure}
\begin{center}
\includegraphics[width=.55\textwidth]{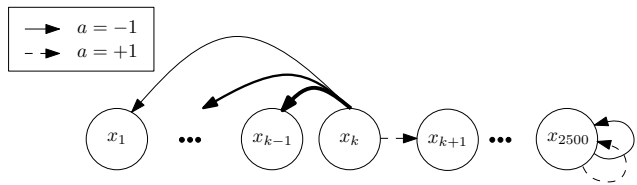}
\end{center}
\caption{
Combination lock: illustration of the combination lock MDP problem.  Nodes
indicate states. State $x_{2500}$ is the goal (absorbing) state and state $x_k$
is an example of interior state.  Arrows indicate possible transitions of 
\textbf{these two nodes only}.  From $x_k$  any previous state is reachable with transition
probability (arrow thickness) proportional to the inverse of the distance to
$x_k$. Among the future states only $x_{k+1}$ is reachable (arrow dashed).
}
\label{fig:combclock}
\end{figure}

\item[Grid world:]
%
%
%
%
%
this MDP consists of a grid of $50\times50$ states. A set of four actions
\{RIGHT, UP, DOWN, LEFT\} is assigned to every state $\inset{x}{X}$. The
location of each state $x$ of the grid is determined by the coordinates
$c_x=(h_x,v_x)$, where  $h_x$ and $v_x$ are some integers between $1$
and $50$. There are $196$ absorbing \emph{firewall states} surrounding the grid
and another one at the center of grid, for which a reward $-1$ is assigned. The
reward for the firewalls is
\begin{equation*}
r(x,a)=-\dfrac{1}{\Norm[2]{c_x}}, \quad\quad\forall\inset{a}{A}.
\end{equation*}
Also, we assign reward $0$ to all of the remaining (non-absorbing) states.

This means that both the top-left absorbing state and the central state have
the least possible reward ($-1$), and that the remaining absorbing states have
reward which increases proportionally to the distance to the state in the
bottom-right corner (but are always negative).

The transition probabilities are defined in the following way: taking action
$a$ from any non-absorbing state $x$ results in a one-step transition in
the direction of action $a$ with probability $0.6$, and a random move to a
state $y\neq x$ with probability inversely proportional to their Euclidean distance
$1/\Norm[2]{c_x-c_y}$. 

The optimal policy then is to \emph{survive} in the grid as long as possible
by avoiding both the absorbing firewalls and the center of the grid.  Note that
because of the difference between the cost of firewalls, the optimal  control prefers the states
near the bottom-right corner of the grid, thus avoiding absorbing states with
higher cost.

\end{description}

\subsubsection{Experimental Setup and Results}
We describe now our experimental setting.  The convergence properties of DPP-RL are 
compared with two other algorithms: a synchronous variant of
Q-learning~\citep{Even-DarM03} (QL), which, like DPP-RL, updates the action-value function 
of all  state-action pairs at each iteration, and the model-based Q-value iteration (VI)
of ~\citet{Kearns99}. VI is a batch reinforcement learning algorithm that first
estimates the model using the whole data set and then performs value iteration on the learned model.

All algorithms are evaluated in terms of $\ell_{\infty}$-norm performance loss
of the action-value function $\Norm{Q^*-Q^{\pi_k}}$ obtained by policy $\pi_k$
induced at iteration $k$. We choose this performance measure in order  to be 
consistent  with the performance measure used in Section~\ref{Approx}. 
The discount factor  $\gamma$ is fixed to $0.995$ and the optimal action-value function
 $Q^*$ is computed with high accuracy through value iteration.

We consider QL with polynomial learning step
$\alpha_k=1/(k+1){^\omega}$ where $\omega\in\{0.51,0.75\}$ and the linear learning step
$\alpha_k=1/(k+1)$.
Note that $\omega$ needs to be larger than $0.5$, otherwise  QL can asymptotically diverge
\citep[see][for the proof]{Even-DarM03}. 

To achieve the best rate of convergence for DPP-RL, we fix $\eta$ to $+\infty$
(see Section~\ref{DPPProof}). This replaces the soft-max operator $\M$ in the
DPP-RL update rule with the max operator $\M[]$, resulting in a greedy policy
$\pi_k$.

To have a fair comparison of  
the three algorithms, since each algorithm requires different number of computations per iteration, we  fix the total computational
budget of the algorithms to the same value for each benchmark.  The computation time is constrained to
$30$ seconds in the case of linear MDP and the combination lock problems.  For
the grid world, which has twice as many actions as the other benchmarks,
the maximum run time is fixed to $60$ seconds.   We also  fix the total number of samples,  per state-action,
 to $1\times10^5$ samples for all problems and algorithms.  Significantly less
number of samples leads to a dramatic decrease of the quality of the obtained
solutions using all the approaches.

 \begin{figure}[t]
\begin{center}
\includegraphics[width=\textwidth]{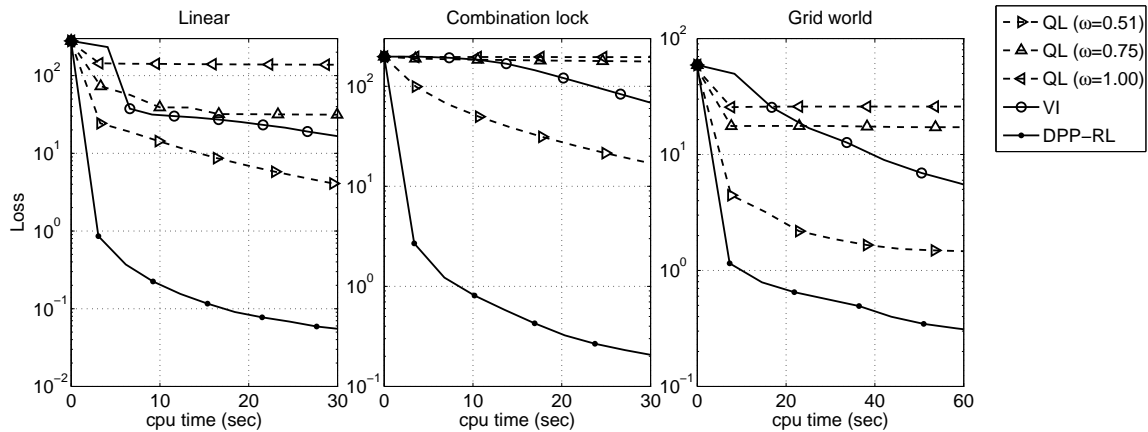}
\end{center}
\caption{
A comparison between DPP-RL, Q-Learning and model-based VI. Each plot
compares the performance loss of the policy induced  by the algorithms for a
different MDP averaged over $50$ different runs (see the text for details). 
}
\label{fig:DPP-RL}
\end{figure}

Algorithms were implemented as MEX files (in C++) and ran on a Intel core i5 processor with 8 GB of memory. cpu time was acquired using the system function \texttt{times()} which provides process-specific cpu time. Randomization was implemented using \texttt{gsl\textunderscore{rng}{\textunderscore}uniform()} function of the GSL library, which is superior to the standard \texttt{rand()}.\footnote{http://www.gnu.org/s/gsl.} Sampling time, which is the same for all algorithms, were not included in cpu time.

Figure \ref{fig:DPP-RL} shows the performance-loss in terms of elapsed cpu time
for the three problems and algorithms.  The results are averages over $50$
runs, where at the beginning of each run~\textbf{(i)}~the action-value function
and the action preferences are randomly initialized in the interval
$[-V_{\max},V_{\max}]$, and~\textbf{(ii)} a new set of samples is generated
from $P(\cdot|x,a)$ for all $\inset{(x,a)}{Z}$.  Results correspond to the
average error computed after a small fixed amount of iterations.

First, we see that DPP-RL converges very fast achieving near optimal
performance after a few seconds.  DPP-RL outperforms both QL and VI in all the
three benchmarks.  The minimum and maximum errors are attained for the linear
MDP problem and the Grid world, respectively. We also observe that the
difference between DPP-RL and QL is very significant, about two orders of
magnitude, in both the linear MDP and the Combination lock problems.  In the
grid world DPP-RL's performance is more than $4$ times better than that of QL.

QL shows the best performance for $\omega=0.51$.  The quality of the QL
solution degrades as a function of $\omega$.  Concerning VI, its error
shows a sudden decrease on the first error caused by the model estimation.

The standard deviations of the performance-loss give an indication of how robust are the
solutions obtained by the algorithms.  Table~\ref{CompSQL}
shows the final numerical outcomes of DPP-RL, QL and VI (standard deviations between
parenthesis). We can see that the variance of estimation of DPP-RL is substantially smaller
than those of QL and VI.
\begin{table}[h]
\caption{A Comparison between  DPP-RL, Q-learning (QL) and the model-based value iteration (VI) given a fixed computational and sampling budget. Table~\ref{CompSQL} shows  error means and standard deviations (between parenthesis) at the end of the simulations for three different algorithms (columns) and three different benchmarks (rows)}
\label{CompSQL}
\centering
\begin{tabular}{|lc|r@{.}lr@{.}l|r@{.}lr@{.}l|r@{.}lr@{.}l|}
\hline
Benchmark&&\multicolumn{4}{c}{Linear MDP} & \multicolumn{4}{c}{Combination lock}&\multicolumn{4}{c|}{Grid world}
\\
\hline
Run Time&&\multicolumn{4}{c}{30 sec.}&\multicolumn{4}{c}{30 sec.}&\multicolumn{4}{c|}{60 sec.}
\\
\hline
\hline
DPP-RL&&$\mathbf{0}$ & $\mathbf{05}$ & $\mathbf{(0}$ & $\mathbf{02)}$ & $\mathbf{0}$ & $\mathbf{20}$ & $\mathbf{(0}$ & $\mathbf{09)}$ & $\mathbf{0}$ & $\mathbf{32}$ & $\mathbf{(0}$ & $\mathbf{03)}$
\\
\hline
VI&& $16$ & $60$ & $(11$ & $60)$ & $69$ & $33$ & $(15$ & $38)$ & $5$ & $67$ & $(1$ & $73)$
\\
   \hline
   \multirow{3}{*}{QL} & 
   $\omega=0.51$   & $4$ & $08$ & $(3$ & $21)$     &   $18$ & $18$ & $(4$ & $36)$   &  $1$ & $46$ & $(0$ & $12)$
   \\
   &$\omega=0.75$  & $31$ & $41$ & $(12$ & $77)$   &   $176$ & $13$ & $(25$ & $68)$ &   $17$ & $21$ & $(7$ & $31)$
   \\
   &$\omega=1.00$  & $138$ & $01$ & $(146$ & $28)$ &   $195$ & $74$ & $(5$ & $73)$  &   $25$ & $92$ & $(20$ & $13)$
   \\
\hline
\end{tabular}
\end{table}

These results show that, as suggested in Theorem~\ref{DPP_RLConv} and~\ref{ADPPMain1}, DPP-RL
manages to average out the simulation noise caused by sampling and converges, rapidly, to a near optimal solution,
 which is very robust.  In addition, we
can conclude that DPP-RL performs significantly better than QL and VI in the
three presented benchmarks for our choice of experimental setup.

\subsection{SADPP}
\label{NumResSADPP}
In this subsection, we illustrate the performance of the SADPP algorithm in the
presence of function approximation and limited sampling budget per iteration.
We compare SADPP with a modification of regularized fitted $Q$-iteration
(RFQI) \citep{FarahmandGSM08} which make use of  a fixed number of basis functions.  RFQI
can be regarded as a Monte-Carlo sampling implementation of approximate value
iteration with action-state representation. We compare   SADPP with RFQI  since both methods make
 use of $\ell_2$-regularization. The purpose of this subsection is
to analyze numerically the sample complexity, i.e, the number of samples required to achieve a near optimal performance with a low variance,
 of SADPP.  The benchmark we
consider is a variant of the \emph{optimal replacement problem} presented in
\citet{Munos08}.  In the following subsection we describe the problem and
subsequently we present the results.

\subsubsection{Optimal replacement problem}
This problem is an infinite-horizon, discounted MDP.  The state measures the
accumulated use of a certain product and is represented as a continuous,
one-dimensional variable.  At each time-step $t$, either the product is kept
$a(t) = 0$ or replaced $a(t) = 1$.  Whenever the product is replaced by a
new one, the state variable is reset to zero $x(t) = 0$, at an additional cost
$C$.  The new state is chosen according to an exponential distribution, with
possible values starting from zero or from the current state value, depending
on the latest action:
\begin{align*}
p(y|x,a=0)&=
    \begin{cases}
        \beta e^{\beta(y-x)} & \text{if $y\geq x$} \\
        0                   & \text{if $y<0$}
    \end{cases}
    &
p(y|x,a=1)&=
    \begin{cases}
        \beta e^\beta y     & \text{if $y\geq 0$} \\
        0                   & \text{if $y<0$}
    \end{cases}.
\end{align*}

The reward function is a monotonically increasing function of the state $x$
if the product is kept $r(x,0)=-c(x)$ and constant if the product
is replaced $r(x,1) = -C-c(0)$.

The optimal action is to keep as long as the accumulated use is below a
threshold or to replace otherwise:
\begin{align*}
a^*(x) &= \begin{cases}
        0 & \text{if $x\in[0,\bar{x}]$}\\
        1 & \text{if $x>\bar{x}$}
    \end{cases}.
\end{align*}

Following \citet{Munos08}, $\bar{x}$ can be obtained exactly via the Bellman
equation and is the unique solution to
\begin{align*} C &= \int_0^{\bar{x}}{
\frac{c'(y)}{1-\gamma}\left(1-\gamma e^{-\beta(1-\gamma)y}\right)dy}.
\end{align*}

\subsubsection{Experimental setup and results}
For both SADPP and RFQI we map the state-action space using $20$ radial basis
functions ($10$ for the continuous one-dimensional state variable $x$, spanning the state space $\mathscr{X}$, and $2$
for the two possible actions).  
Other parameter values where chosen
to be the same as in \citet{Munos08}, that is, $\gamma=0.6, \beta=0.5, C=30$
and $c(x)=4x$, which results in $\bar{x}\simeq 4.8665$.  We also fix an upper
bound for the states, $x_{\max} = 10$ and modify the problem definition such
that if the next state $y$ happens to be outside of the domain $[0, x_{\max}]$
then the product is replaced immediately, and a new state is drawn as if
action $a=1$ were chosen in the previous time step.

To compare both Algorithms we discretize the state space in $K=100$ bins
and use the following error measure:
\begin{align}
\label{eq:err}
\text{error} &=\frac{\sum_{k=1}^{K}{|a^*(x_k) - \hat{a}(x_k)|}}{K},
\end{align}
where $\hat{a}$ is the action selected by the Algorithm.  Note
that, unlike RFQI which selects the  action by choosing the
action with the highest action-value function, SADPP induces a stochastic
policy, that is, a distribution over actions.  We select $\hat{a}$ for SADPP by choosing
the most probable action from the induced soft-max policy, and then use
this to compute \eqref{eq:err}. Both algorithms were implemented  in MatLab  
and executed under the same hardware specifications of the previous section.

We analyze the effect of using different number of samples $N$ per iteration,
$N\in\{50,150,500\}$.\footnote{For both algorithms a new independent set of samples are generated
at each iteration.} The results are averages over $200$
runs, where at the beginning of each run the vector $\theta$ is initialized in the interval
$[-1,1]$ for both algorithms. The rest of the parameters, including the regularization factor $\alpha$ and $\eta$,
were optimized for the best asymptotic performance for each $N$ independently.
  
\begin{figure}
\begin{center}
\includegraphics[width=\textwidth]{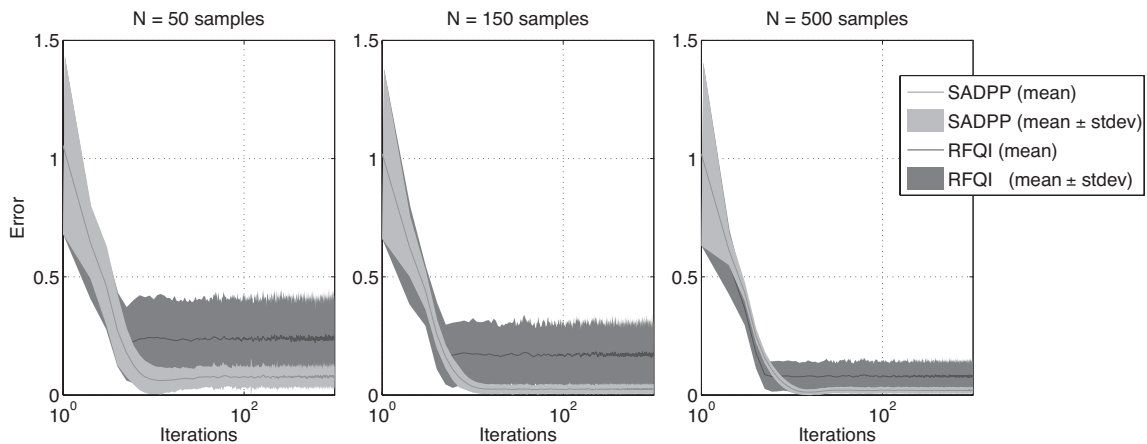}
\end{center}
\caption{Numerical results for the optimal replacement problem. Each plot shows
a comparison of the error between SDAPP and RFQI Algorithms using certain
number of samples $N$.  Error is defined as in equation \eqref{eq:err}.  Areas
indicate averages plus/minus standard deviations of the error starting from
$200$ uniformly distributed initial conditions (see the text for details).}
\label{fig:repl}
\end{figure}

Figure~\ref{fig:repl} shows averages and standard deviations of the errors.
First, we observe that for large $N$, after an initial transient, both SADPP
and RFQI reach a near optimal solution. We observe that SADPP asymptotically
outperforms RFQI on average in all cases.  The average error and the variance
of estimation of the resulting solutions decreases with $N$ in both
approaches.  A comparison of the variances after the transient suggests that
the sample complexity of SADPP is significantly smaller than RFQI.
Remarkably, the variance of SADPP using $N=50$ samples is comparable to the
one provided by RFQI using $N=500$ samples.  Further, the variance of SADPP
is reduced faster with increasing $N$.  These results allow to conclude that
SADPP can have positive effects in reducing the effect of simulation noise, as
postulated in Section~\ref{Approx}.

%% file: jmlr_draft_arxiv.bbl
\begin{thebibliography}{36}
\providecommand{\natexlab}[1]{#1}
\providecommand{\url}[1]{\texttt{#1}}
\expandafter\ifx\csname urlstyle\endcsname\relax
  \providecommand{\doi}[1]{doi: #1}\else
  \providecommand{\doi}{doi: \begingroup \urlstyle{rm}\Url}\fi

\bibitem[Antos et~al.(2008)Antos, Munos, and Szepesv{\'a}ri]{Antos07}
A.~Antos, R.~Munos, and Cs. Szepesv{\'a}ri.
\newblock Fitted {Q}-iteration in continuous action-space {MDP}s.
\newblock In \emph{Proceedings of the 21st Annual Conference on Neural
  Information Processing Systems}. MIT Press, 2008.

\bibitem[Bertsekas(2007)]{Berst07b}
D.~P. Bertsekas.
\newblock \emph{Dynamic Programming and Optimal Control}, volume~II.
\newblock Athena Scientific, third edition, 2007.

\bibitem[Bertsekas and Tsitsiklis(1996)]{Tsitsiklis96}
D.~P. Bertsekas and J.~N. Tsitsiklis.
\newblock \emph{Neuro-Dynamic Programming}.
\newblock Athena Scientific, 1996.

\bibitem[Bhatnagar et~al.(2009)Bhatnagar, Sutton, Ghavamzadeh, and
  Lee]{BhatnagarSGL09}
S.~Bhatnagar, R.~S. Sutton, M.~Ghavamzadeh, and M.~Lee.
\newblock Natural actor-critic algorithms.
\newblock \emph{Automatica}, 45\penalty0 (11):\penalty0 2471--2482, 2009.

\bibitem[de~Farias and Roy(2000)]{Farias00}
D.~P. de~Farias and B.~Van Roy.
\newblock On the existence of fixed points for approximate value iteration and
  temporal-difference learning.
\newblock \emph{Journal of Optimization Theory and Applications}, 105\penalty0
  (3):\penalty0 589--608, 2000.

\bibitem[Even-Dar and Mansour(2003)]{Even-DarM03}
E.~Even-Dar and Y.~Mansour.
\newblock Learning rates for {Q}-learning.
\newblock \emph{Journal of Machine Learning Research}, 5:\penalty0 1--25, 2003.

\bibitem[Farahmand et~al.(2008)Farahmand, Ghavamzadeh, Szepesv{\'a}ri, and
  Mannor]{FarahmandGSM08}
A.~Farahmand, M.~Ghavamzadeh, Cs. Szepesv{\'a}ri, and S.~Mannor.
\newblock Regularized fitted {Q}-iteration: Application to planning.
\newblock In \emph{European Workshop on Reinforcement Learning}, Lecture Notes
  in Computer Science. Springer, 2008.

\bibitem[Farahmand et~al.(2010)Farahmand, Munos, and
  Szepesv{\'a}ri]{AmirRemi10}
A.~Farahmand, R.~Munos, and Cs. Szepesv{\'a}ri.
\newblock Error propagation for approximate policy and value iteration.
\newblock In \emph{Proceedings of the 23rd Annual Conference on Neural
  Information Processing Systems}. MIT Press, 2010.

\bibitem[Hoffmann-J{\o}rgensen and Pisier(1976)]{Hoffman76}
J.~Hoffmann-J{\o}rgensen and G.~Pisier.
\newblock The law of large numbers and the central limit theorem in banach
  spaces.
\newblock \emph{The Annals of Probability}, 4\penalty0 (4):\penalty0 587--599,
  1976.

\bibitem[Jaakkola et~al.(1994)Jaakkola, Jordan, and Singh]{Jaakkola94}
T.~Jaakkola, M.~I. Jordan, and S.~Singh.
\newblock On the convergence of stochastic iterative dynamic programming.
\newblock \emph{Neural Computation}, 6\penalty0 (6):\penalty0 1185--1201, 1994.

\bibitem[Kakade(2001)]{Kakade02}
S.~Kakade.
\newblock Natural policy gradient.
\newblock In \emph{Advances in Neural Information Processing Systems 14}. MIT
  Press, 2001.

\bibitem[Kappen(2005)]{Kappen05b}
H.~J. Kappen.
\newblock Path integrals and symmetry breaking for optimal control theory.
\newblock \emph{Statistical Mechanics}, 2005\penalty0 (11):\penalty0 P11011,
  2005.

\bibitem[Kearns and Singh(1999)]{Kearns99}
M.~Kearns and S.~Singh.
\newblock Finite-sample convergence rates for {Q}-learning and indirect
  algorithms.
\newblock In \emph{Advances in Neural Information Processing Systems 12}. MIT
  Press, 1999.

\bibitem[Kober and Peters(2008)]{KoberP08}
J.~Kober and J.~Peters.
\newblock Policy search for motor primitives in robotics.
\newblock In \emph{Proceedings of the 21st Annual Conference on Neural
  Information Processing Systems}. MIT Press, 2008.

\bibitem[Koenig and Simmons(1993)]{Koenig93}
Sven Koenig and Reid~G. Simmons.
\newblock Complexity analysis of real-time reinforcement learning.
\newblock In \emph{Proceedings of the Eleventh National Conference on
  Artificial Intelligence}. AAAI Press, 1993.

\bibitem[Konda and Tsitsiklis(2003)]{Konda03}
V.~Konda and J.~N. Tsitsiklis.
\newblock On actor-critic algorithms.
\newblock \emph{SIAM Journal on Control and Optimization}, 42\penalty0
  (4):\penalty0 1143--1166, 2003.

\bibitem[Lagoudakis and Parr(2003)]{Lago03}
M.~G. Lagoudakis and R.~Parr.
\newblock Least-squares policy iteration.
\newblock \emph{Journal of Machine Learning Research}, 4:\penalty0 1107--1149,
  2003.

\bibitem[MacKay(2003)]{MacKay03}
D.~J.~C. MacKay.
\newblock \emph{Information Theory, Inference, and Learning Algorithms}.
\newblock Cambridge University Press, 2003.

\bibitem[Maei et~al.(2010)Maei, Szepesv{\'a}ri, Bhatnagar, and
  Sutton]{MaeiSBS10}
H.~Maei, Cs. Szepesv{\'a}ri, S.~Bhatnagar, and R.~S. Sutton.
\newblock Toward off-policy learning control with function approximation.
\newblock In \emph{Proceedings of the 27th Annual International Conference on
  Machine Learning.} Omnipress, 2010.

\bibitem[Melo et~al.(2008)Melo, Meyn, and Ribeiro]{Melo08}
F.~Melo, S.~Meyn, and I.~Ribeiro.
\newblock An analysis of reinforcement learning with function approximation.
\newblock In \emph{Proceedings of 25 International Conference on Machine
  Learning}. ACM, 2008.

\bibitem[Munos(2005)]{Remi05}
R.~Munos.
\newblock Error bounds for approximate value iteration.
\newblock In \emph{Proceedings of the 20th National Conference on Artificial
  Intelligence}, volume~II. AAAI Press, 2005.

\bibitem[Munos and Szepesv{\'a}ri(2008)]{Munos08}
R.~Munos and Cs. Szepesv{\'a}ri.
\newblock Finite-time bounds for fitted value iteration.
\newblock \emph{Journal of Machine Learning Research}, 9:\penalty0 815--857,
  2008.

\bibitem[Perkins and Precup(2002)]{PerkinsP02}
T.~J. Perkins and D.~Precup.
\newblock A convergent form of approximate policy iteration.
\newblock In \emph{Advances in Neural Information Processing Systems 15}. MIT
  Press, 2002.

\bibitem[Peters and Schaal(2008)]{Peters08}
J.~Peters and S.~Schaal.
\newblock Natural actor-critic.
\newblock \emph{Neurocomputing}, 71\penalty0 (7--9):\penalty0 1180--1190, 2008.

\bibitem[Peters et~al.(2010)Peters, M{\"u}lling, and Altun]{PetersMA10}
J.~Peters, K.~M{\"u}lling, and Y.~Altun.
\newblock Relative entropy policy search.
\newblock In \emph{Proceedings of the Twenty-Fourth AAAI Conference on
  Artificial Intelligence}. AAAI Press, 2010.

\bibitem[Singh et~al.(2000)Singh, Jaakkola, Littman, and Szepesvari]{singh00}
S.~Singh, T.~Jaakkola, M.L. Littman, and Cs. Szepesvari.
\newblock Convergence results for single-step on-policy reinforcement-learning
  algorithms.
\newblock \emph{Machine Learning}, 38\penalty0 (3):\penalty0 287--308, 2000.

\bibitem[Strehl et~al.(2009)Strehl, Li, and Littman]{StrehlLL09}
A.~L. Strehl, L.~Li, and M.~L. Littman.
\newblock Reinforcement learning in finite {MDP}s: {PAC} analysis.
\newblock \emph{Journal of Machine Learning Research}, 10:\penalty0 2413--2444,
  2009.

\bibitem[Sutton and Barto(1998)]{Sutton98}
R.~S. Sutton and A.~G. Barto.
\newblock \emph{Reinforcement Learning: An Introduction}.
\newblock MIT Press, 1998.

\bibitem[Sutton et~al.(1999)Sutton, McAllester, Singh, and Mansour]{Sutton00}
R.~S. Sutton, D.~McAllester, S.~Singh, and Y.~Mansour.
\newblock Policy gradient methods for reinforcement learning with function
  approximation.
\newblock In \emph{Advances in Neural Information Processing Systems 12}. MIT
  Press, 1999.

\bibitem[Szepesv{\'a}ri(1997)]{Szepesvari97}
Cs. Szepesv{\'a}ri.
\newblock The asymptotic convergence-rate of {Q}-learning.
\newblock In \emph{Advances in Neural Information Processing Systems 10}. MIT
  Press, 1997.

\bibitem[Szepesv{\'a}ri(2010)]{Szepesvari2010}
Cs. Szepesv{\'a}ri.
\newblock \emph{Algorithms for Reinforcement Learning}.
\newblock Synthesis Lectures on Artificial Intelligence and Machine Learning.
  Morgan {\&} Claypool Publishers, 2010.

\bibitem[Szepesv{\'a}ri and Smart(2004)]{Szepesva04}
Cs. Szepesv{\'a}ri and W.~Smart.
\newblock Interpolation-based q-learning.
\newblock In \emph{Proceedings of 21st International Conference on Machine
  Learning}, ACM, 2004.

\bibitem[Thiery and Scherrer(2010)]{Thiery10}
C.~Thiery and B.~Scherrer.
\newblock Least-squares lambda policy iteration: Bias-variance trade-off in
  control problems.
\newblock In \emph{Proceedings of the 27th Annual International Conference on
  Machine Learning.} Omnipress, 2010.

\bibitem[Todorov(2006)]{Todorov06}
E.~Todorov.
\newblock Linearly-solvable {M}arkov decision problems.
\newblock In \emph{Proceedings of the 20th Annual Conference on Neural
  Information Processing Systems}. MIT Press, 2006.

\bibitem[Vlassis and Toussaint(2009)]{VlassisT09}
N.~Vlassis and M.~Toussaint.
\newblock Model-free reinforcement learning as mixture learning.
\newblock In \emph{Proceedings of the 26th Annual International Conference on
  Machine Learning.} ACM, 2009.

\bibitem[Watkins and Dayan(1992)]{Watkins92}
C.J.C.H. Watkins and P.~Dayan.
\newblock Q-learning.
\newblock \emph{Machine Learning}, 3\penalty0 (8):\penalty0 279--292, 1992.

\end{thebibliography}
